\documentclass{article}
\usepackage{PRIMEarxiv}

\usepackage{longtable}

\usepackage[utf8]{inputenc} 
\usepackage[T1]{fontenc}    
\usepackage{hyperref}       
\usepackage{url}            
\usepackage{booktabs}       
\usepackage{amsfonts}       
\usepackage{nicefrac}       
\usepackage{microtype}      
\usepackage{lipsum}         
\usepackage{fancyhdr}       
\usepackage{graphicx}       
\usepackage{graphbox}       
\graphicspath{{media/}}     
\usepackage{appendix}       %
\usepackage{color}          %
\usepackage{multirow}       %
\usepackage{booktabs}       %
\usepackage{subcaption}     %
\usepackage{soul}           %
\usepackage{makecell}
\usepackage{orcidlink}
\usepackage{xcolor}

\newcounter{ModelCounter}
\newcommand{\modellabel}[1]{\refstepcounter{ModelCounter}\texttt{M\arabic{ModelCounter}}\label{#1}}
\newcommand{\modelref}[1]{\hyperref[#1]{\texttt{M\ref*{#1}}}}

\title{\normalfont Learning to See More: UAS-Guided Super-Resolution of Satellite Imagery for Precision Agriculture}

\author{
  Arif Masrur\,\orcidlink{0000-0001-5050-407X}\thanks{Corresponding author: \href{mailto:amasrur.du@gmail.com}{amasrur.du@gmail.com} (or \href{mailto:amasrur@esri.com}{amasrur@esri.com})}\\
  Esri \\
  New York, NY \\
   \And
  Peder A. Olsen\,\orcidlink{0000-0002-3836-8017}\\
  Microsoft Research \\
  Redmond, WA\\
     \And
  Paul R. Adler\,\orcidlink{0000-0002-6787-631X} \\ 
  USDA - Agricultural Research Service \\
  University Park, PA\\
       \And
  Carlan Jackson\\
  Dept. of Electrical Engineering and Computer Science\\
  Alabama A\&M University, AL \\
       \And
  Matthew W. Myers \\
  USDA - Agricultural Research Service \\
  University Park, PA\\
         \And
  Nathan Sedghi\,\orcidlink{0000-0003-3935-3690} \\
  Dept. of Environmental Science and Technology \\
  Univ. of Maryland, College Park, MD\\
           \And
  Ray R. Weil\,\orcidlink{0000-0001-9658-7966}\\
  Dept. of Environmental Science and Technology \\
  Univ. of Maryland, College Park, MD\\
}

\begin{document}
\maketitle

\begin{abstract}
Unmanned Aircraft Systems (UAS) and satellites are important data sources for precision agriculture, yet each presents trade-offs.  Satellite data provide broad temporal, spatial and spectral coverage but lack the fine resolution needed for many precision farming applications, while UAS can provide high spatial detail but are limited by coverage and cost constraints - especially for hyperspectral data. This study introduces a novel framework that fuses satellite and UAS imagery using neural network-based super-resolution methods. By integrating data across the spatial, spectral, and temporal domains, we can leverage the strengths of both platforms in a cost-effective manner. We used estimation of cover crop biomass and nitrogen (N) as a case study to evaluate our methodology. By spectrally extending the UAS RGB data to the critical vegetation red edge and near-infrared regions, we created high-resolution Sentinel-2 imagery and improved the accuracy of biomass and N estimation by 18\% and 31\%, respectively.  Importantly, our results demonstrate that UAS data need only be collected from a subset of fields and time points.  From these limited observations, farmers can then 1) enhance the spectral detail of UAS RGB imagery; 2) increase the spatial resolution by using satellite data; and 3) extend these enhancements spatially and across the growing season at the frequency of the satellite flights.  Our SRCNN-based \textit{spectral extension model} showed considerable promise for model transferability over other cropping systems in the Upper and Lower Chesapeake Bay regions.  Additionally, it remains effective even when cloud-free satellite data are unavailable, relying solely on the UAS RGB input.  The \textit{spatial extension model} produces better biomass and N predictions than models built on raw UAS RGB images.  Thus, \textit{the farmer can stop flying UAS} once a specialized spatial extension model has been trained from targeted UAS RGB data.  While several super-resolution innovations are introduced, the core contribution is a practical, scalable system for precision agriculture. The model is lightweight by design to ensure accessibility and affordability for on-farm use.
\end{abstract}


\section{Introduction}
Remote sensing (RS) technologies provide tremendous opportunities to improve precision crop management decisions \cite{seelan2003remote, weiss2020remote}. While the effectiveness of these technologies improves with higher spectral range and spatial resolution \cite{hunt2018good}, these characteristics vary widely across platforms. Unmanned Aircraft Systems (UAS) have become indispensable tools in precision agriculture \cite{radoglou2020compilation, singh2020hyperspectral, zhang2012application},  enabling timely, high-resolution field-level assessments of crop biomass \cite{wang2021applications} and nitrogen (N) status \cite{argento2021site, gruner2021prediction}, as well as detection of weeds, diseases, and pest infestations \cite{dash2018uav, watt2017use, zhu2024intelligent}. However, despite their spatial precision, UAS-based imaging is constrained by limited spectral range and coverage due to high operational costs \cite{de2021agriq}. In contrast, freely available satellite imagery offers broad spatial and temporal coverage but often lacks the fine resolution required for precision agriculture applications. To bridge this gap, AI-based image fusion techniques \cite{samadzadegan2025critical}, typically dominated by the pan-sharpening \cite{li2022deep}, have been employed to enhance satellite image resolution. A more effective strategy incorporates high-resolution hyperspectral ground-truth data from UAS or airborne platforms, which better support model development. However, these methods are limited when training data consist of imagery from sensors with differing spatial and spectral characteristics. This creates a two-fold challenge for the fusion models: learn to enhance both spatial resolution and spectral richness simultaneously. To address this, we can utilize \textit{super-resolution} methods \cite{dong2015image} from computer vision to reconstruct high-resolution images from low-resolution inputs by learning spatial detail using deep learning (DL) methods. Typically, in super-resolution for low-resolution remote sensing platforms, the high-resolution training targets have different spectral characteristics. By simulating a low-resolution RS platform (e.g., Sentinel-2) at a higher resolution based on UAS hyperspectral data, we can provide high-resolution training targets that \textit{match the spectral properties} of the low-resolution multispectral sensor, thus allowing the DL model to focus on the super-resolution task only, removing the spectral distortion issue entirely. In this study, we introduce a novel framework based on super-resolution and spectral extension to simultaneously enhance the spatial and spectral fidelity of satellite data. This approach offers a scalable and cost-effective solution for both research and operational precision agriculture across a range of spectral, spatial, and temporal domains. 

\begin{figure}[h!]
  \centering
  \includegraphics[width=0.8\textwidth]{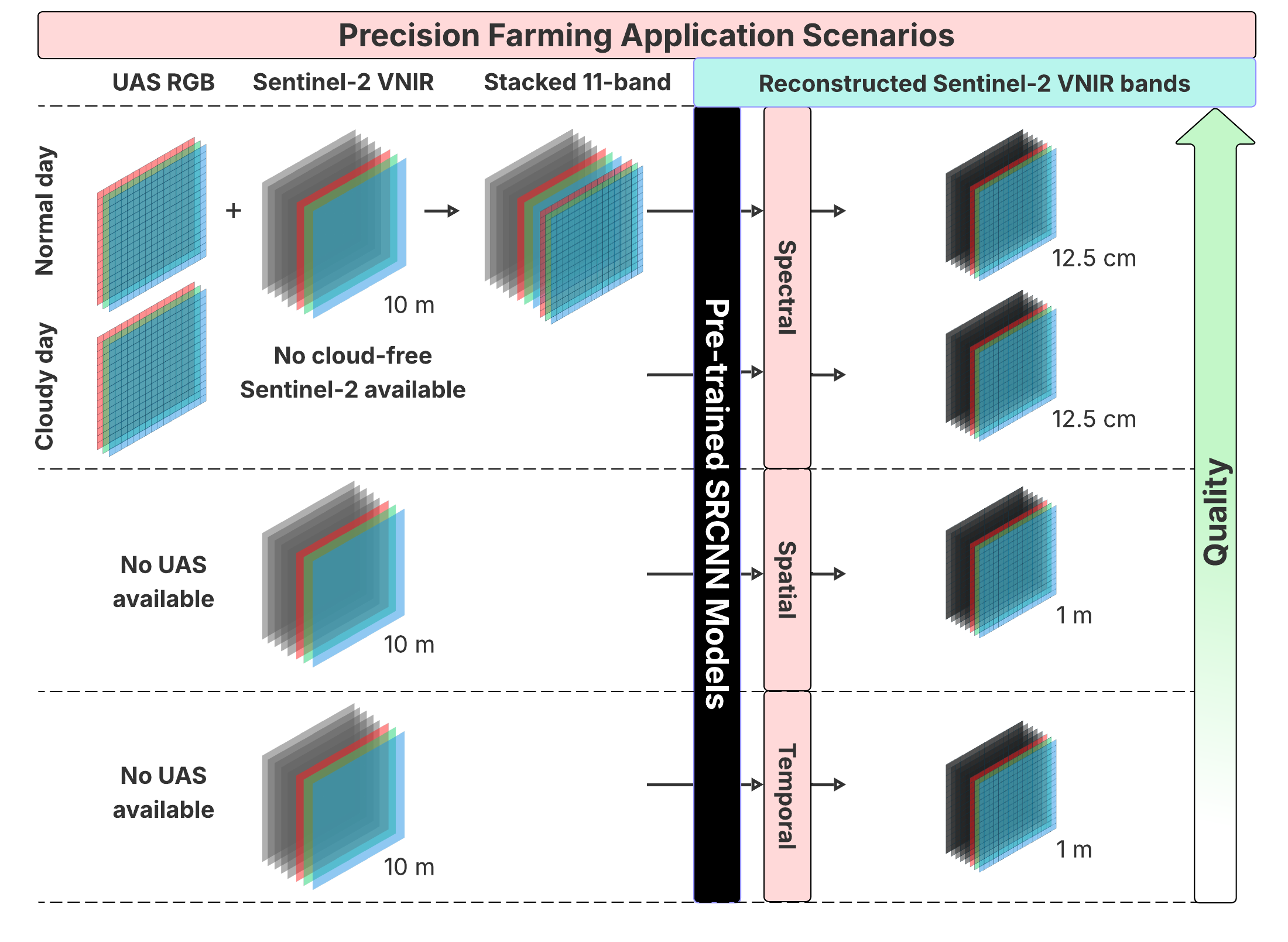}
  \caption{Application scenarios of super-resolution in cost-effective precision farming. In all these scenarios we present neural networks that improved the performance over existing methods using original UAS RGB or Sentinel-2 datasets (see Tables \ref{tab:tab2} and \ref{tab:cloudy}).}
  \label{fig:application}
\end{figure}

In practical agricultural applications, the choice of an appropriate RS platform is influenced by factors such as availability, cost, and suitability for specific tasks. Figure~\ref{fig:application} illustrates how our super-resolution framework can be utilized in various scenarios by leveraging existing RS platforms like Sentinel-2 and UAS to improve upon traditional methods. We observe consistent qualitative and quantitative improvements in image resolution and spectral information content across all these scenarios--whether UAS RGB, Sentinel-2, or both are available. In at least one scenario demonstrated in this study, our method enables farmers to eliminate the need for UAS flights altogether, achieving comparable performance using only satellite-based imagery. 

Our model that fuses UAS and satellite data -- through a process we term as spectral extension -- produces a very high-fidelity image reconstructions at sub-meter spatial resolution. Combining UAS RGB with satellite imagery in this way unlocks access to critical remote sensing indices, such as those based on vegetation red edge (VRE) and near-infrared (NIR) bands, which are not available from RGB sensors alone.  An early version of this spectral extension model is available via the FarmVibes open source repository\footnote{\href{https://github.com/microsoft/farmvibes-ai/blob/main/notebooks/spectral_extension/spectral_extension.ipynb}{https://github.com/microsoft/farmvibes-ai/blob/main/notebooks/spectral\_extension/spectral\_extension.ipynb}}.
Additionally, these spectrally enriched high-resolution outputs can then be repurposed to train new satellite-only super-resolution models that produce sharpened images over the non-flown areas. We term this process as spatial extension, which enables the development of crop- and field-specific supervised learning models without the need for ground-truth data from UAS RGB or costly hyperspectral sensors. To better contextualize our super-resolution modeling approach and its contributions, we next review the historical development and current best practices in the use of UAS and satellite-based remote sensing for precision agriculture. 

\subsection{The historical context}
Historically, both satellite and UAS platforms have been used to estimate key characteristics of plants and soil, such as above-ground biomass \cite{schreiber2022above} and nitrogen (N) \cite{belgiu2023prisma, berger2020crop}, with studies showing that higher spatial resolution and broader spectral range are associated with improved estimation accuracy. This suggests that hyperspectral sensors covering the 400–2,500 nm range could provide the most accurate models, but it may not be feasible or cost-effective to fly large acreage with a UAS platform. This trade-off has led to increasing interest in image fusion techniques that combine the spatial detail of UAS with the spectral breadth and coverage of satellite data. Together, they have supported applications ranging from crop monitoring \cite{maimaitijiang2020crop}, physiological stress detection \cite{dash2018uav, sagan2019uav} to forest resource management \cite{marx2017uav, puliti2018combining} and habitat monitoring \cite{stark2018combining}. 

\begin{figure}[ht]
\centering
  \includegraphics[width=0.55\textwidth]{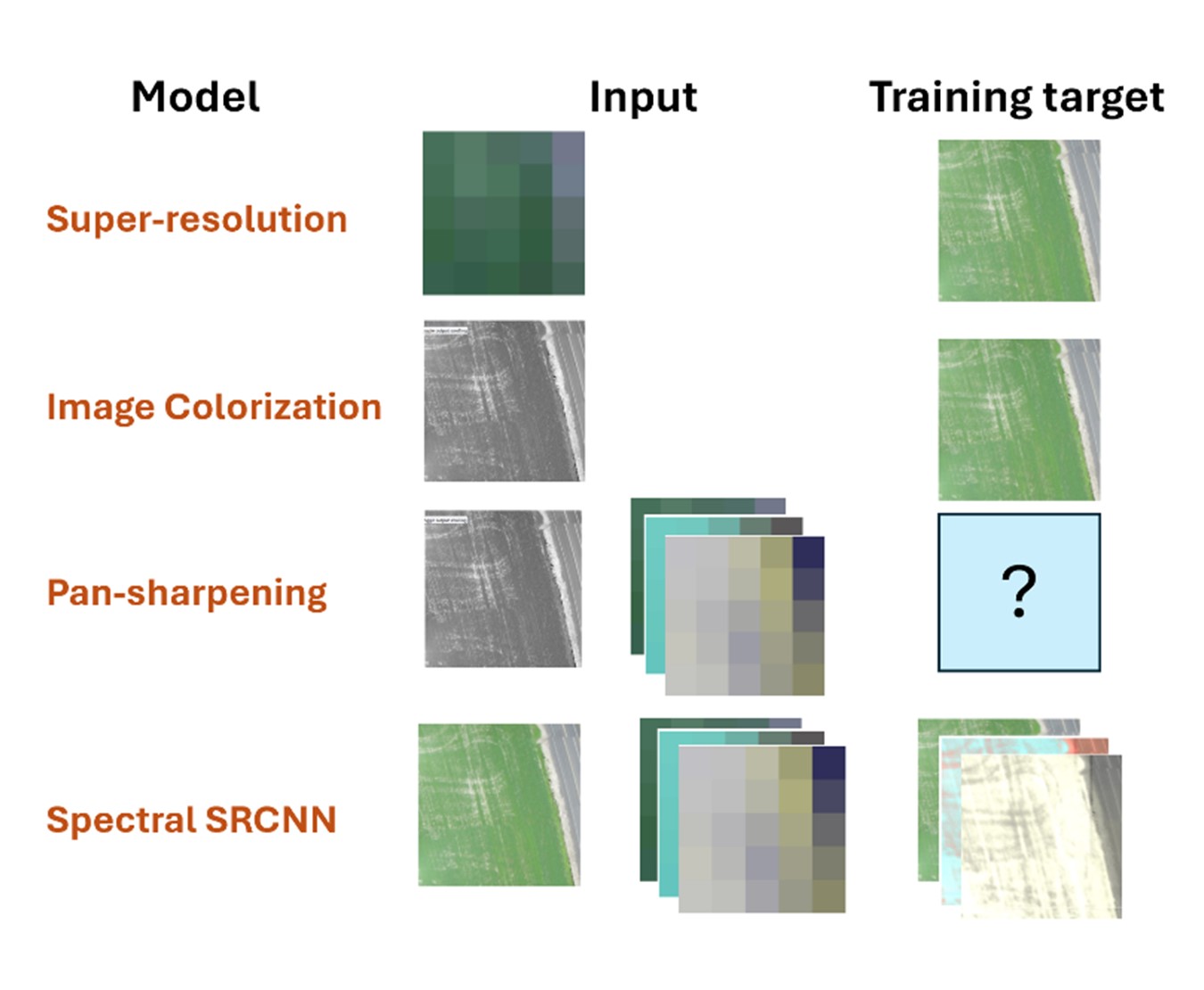}
  \caption{A comparison of common super-resolution modeling frameworks in terms of the input and output targets' structure. The proposed Spectral SRCNN (see Section~\ref{sec:extensions}) can combine frameworks by simulating the satellite sensor at high-resolution using hyperspectral UAS imagery.}
  \label{fig:super-resolution-frameworks}
\end{figure}

This synergy between satellites and UAS is achieved through various image fusion approaches based mainly on methods such as pan-sharpening, decomposition, and deep learning  \cite{samadzadegan2025critical}. Image fusion occurs at the pixel, feature and decision levels \cite{li2017pixel}, with the common objective of improving the spatial resolution of the low-resolution image while preserving broad contextual and spectral information. Many DL-based pan-sharpening methods are applied at the pixel level to fuse multispectral (MS) and panchromatic (Pan) images, hyperspectral (HS) and Pan images, as well as paired HS and MS images that include autoencoder, convolutional neural network (CNN), generative adversarial network (GAN), and visual transformer (ViT) \cite{li2022deep}. Many methods treat image fusion as an image super-resolution problem that recovers a high-resolution image from a given low-resolution one \cite{dong2016accelerating}, thus it allows a DL model to learn the relationship between high-resolution and low-resolution image patches. Super-resolution approaches have been shown to be successful for pixel upscaling factors in the range 2-10 for Sentinel-2 \cite{adigun2022location, galar2020super, lanaras2018super, salgueiro2020super, tarasiewicz2023multitemporal} and historic Landsat imagery \cite{kong2023super}, and are increasingly being applied in precision agriculture \cite{jonak2024spagri, meng2024farmsr}. Image colorization is another technique that augments a gray-scale or single-channel image \cite{wu2021remote}, but may suffer from pixel's intensity mismatch between input and output image, thus a simultaneous super-resolution and colorization is proposed by \cite{liu2018single}. Figure~\ref{fig:super-resolution-frameworks} gives an overview of the different approaches of super-resolution, and it should be noted that pan-sharpening methods are typically unsupervised while traditional super-resolution methods applied to satellite-based imagery are supervised such that they use higher resolution training targets with different spectral characteristics. 

Pixel-level fusion of satellite MS and UAS images is generally a challenging task, even for state-of-the-art super-resolution techniques \cite{toosi2025toward}, because the pixel resolution in the satellite domain is typically much larger (1-10 m) compared to the UAS domain (1-5 cm). One satellite image pixel may easily cover 10,000 or more UAS pixels, requiring an upscaling factor of 8-10 to obtain a high resolution image from a low resolution image. As a result, fusion between UAS and spaceborne images has traditionally occurred at a late stage, where each platform contributes an independent decision or higher-level features that are then combined \cite{alvarez2021uav, ouhami2021computer}. 
Unlike single-modal fusion tasks, such as panchromatic and MS fusion, 
fusion across different platforms presents additional complexity -- particularly in image registration, which aligns the pixels across the images. When source images originate from disparate platforms such as Sentinel-2 and UAS, accurate image-to-image registration becomes critical to fusion quality, according to \cite{li2022deep}. 

\subsection{Spatial, spectral and temporal extension}
\label{sec:extensions}
In our work, we directly address these challenges by developing a comprehensive image fusion framework that integrates spectral alignment, pixel alignment (image registration), and super-resolution strategies. Super-resolution has been shown to be effective when the resolution gap between aligned pixels is within a factor of 8–10. To this end, we build \textit{spatial extension} super-resolution convolutional neural network (SRCNN) models that aggressively reduce the resolution from 10 m to 1 m when only low-resolution satellite data is available. This approach is also applied to a \textit{temporal extension} scenario where the farmer needs to construct high-resolution data for the same field but for a different temporal period that lacks UAS data. Ideally, a temporal extension would be a scenario in which UAS data are collected for one or more years before UAS data collection is stopped. When additional high-resolution data are available, we propose an unconventional super-resolution with a side-information approach. Using UAS RGB imagery co-aligned with Sentinel-2 MS imagery allows us to match our chosen UAS resolution (12.5 cm in our dataset, derived from data originally collected at 3 cm resolution). We refer to this as a \textit{spectral extension} model, which adds spectral richness from the satellite bands (i.e., spectral bands in the 700-900 nm range) not available to an UAS RGB sensor. In this context, we explore several deep learning architectures beyond SRCNN (see Section~\ref{sec:super-resolution-methods}). In scenarios where cloud cover prevents acquisition of high quality co-aligned satellite image scenes, it is also possible to extend the spectral range from the UAS image alone. Such a companion model uses the spatial context to infer the missing spectral data but is not as accurate as when cloud-free satellite imagery is available. 

Training a spectral extension model requires access to high-resolution ground-truth data that match the spectral characteristics of satellite imagery. While pan-sharpening methods \cite{javan2021review}, such as Pan-GAN \cite{ma2020pan}, allow training without explicit ground-truth data by preserving high-frequency features from the panchromatic band (see Figure~\ref{fig:super-resolution-frameworks}), we find that incorporating high-resolution bands (e.g., UAS RGB) as side information and having access to all the satellite bands at high resolution for training targets produces qualitatively better results. UAS-collected hyperspectral imagery is used to generate this high-resolution ground-truth. Specifically, we extracted 17 TB of cloud-free UAS hyperspectral data acquired with the Headwall Nano-Hyperspec sensor, which operates in the VNIR (Visible and Near-Infrared; 400–1,000 nm) range and captures 269-270 spectral bands at a spectral resolution of 6 nm (the bands are spectrally separated by 2.2 nm with a bandwidth around 6 nm - we used 6 nm as a conservative definition of spectral resolution). These data are matched with the corresponding cloud-free Sentinel-2 scenes. To simulate the spectral response of the Sentinel-2 Multispectral Instrument (MSI), we use non-negative linear regression guided by the MSI’s published spectral response functions \cite{COPE-GSEG-EOPG-TN-15-0007}. Importantly, the hyperspectral-derived MSI simulations do not contain the atmospheric noise present in actual spaceborne observations, resulting in cleaner ground-truth. After spectrally harmonizing the sensors, it is equally important to spatially align them by transforming the UAS data to the satellites' coordinate reference system and co-locating image corners to the satellite pixel corners. Following that, image registration is used for further satellite sub-pixel alignment — a particularly critical step when using lower-cost, commodity UAS systems.    

In spatial and temporal extension scenarios, UAS flights can be strategically scheduled to capture representative data throughout the farm and the growing season, enabling the model to adapt to specific crops, sites, and temporal stages. Once trained on a full season’s data, the model generalizes well enough to eliminate the need for continued UAS flights. As we will demonstrate, our spectral extension model generalizes well to other regions and crops. The generated images can serve as high-resolution ground-truth for training spatial extension models tailored to new locations and crops. While this approach may incur a modest performance trade-off compared to models trained directly on hyperspectral data, it significantly reduces operational costs. Thus, farmers can avoid investing in expensive hyperspectral equipment, which can cost as much as \$175,000, while still achieving performance accuracy that meets or surpasses what is possible with UAS RGB imagery alone.

The novelty of this study lies in the development of an end-to-end spatially and temporally scalable system that integrates spectral simulation of low-resolution MS imagery with super-resolution guided by side information to enhance both spatial and spectral resolution, making the use of scale-appropriate remote sensing tools for precision farming more cost-effective and accurate. Although this work builds upon established deep learning methods, it was far from certain that a spectral extension system could be capable of enhancing Sentinel-2 data from 10 m to 12.5 cm. To our knowledge, both the simulation of Sentinel-2 MS data using hyperspectral data from UAS platforms and the use of super-resolution with side information have not been previously demonstrated in the context of precision agriculture. The most comparable prior study \cite{brook2020smart} used a handheld spectrometer to collect sparse ground-truth and utilized a pan-sharpening-inspired modeling approach, lacking the spatio-temporal scalability and spectral enhancements offered by our framework. Another single platform fusion study by \cite{lanaras2018super} used a deep CNN-based super-resolution method to upsample Sentinel-2’s 20 m and 60 m bands to 10 m resolution (2× and 6×). They accomplished this by building a super-resolution network for 40 to 20 m upsampling (360 to 60 m for the 60m bands) and applying it to super-resolve 20 m input to 10 m.  They show that data mismatch causes the RMSE to increase by 50\%, yet their methodology still surpasses that of the best pan-sharpening methods. (This result is reasonable as classical pan-sharpening do not have ground truth training data). We go one step further in our study to provide ground truth data at the target resolution by aligning high-resolution hyperspectral data that are used to simulate the ground truth. Moreover, our framework aims to achieve far greater upsampling—from 10 m to 12.5 cm when high-resolution UAS RGB data are available.  Therefore, beyond intra-Sentinel-2 resolution harmonization, our method integrates heterogeneous data sources to enable both spectral and spatio-temporal extensions to support generalization across crops, regions, and time periods for precision agriculture. 

\subsection{Study outline}
To demonstrate the utility of our proposed super-resolution framework, we apply it to the improved monitoring of cover crop health characterized by the status of biomass yield and nitrogen (N) content. Winter cover cropping is an important component of sustainable agriculture in the Upper and Lower Chesapeake Bay regions. Accurately measuring and optimizing biomass yield and N content play critical role in precision farming to support improved nutrient management, soil health, and the sustainability of cropping systems \cite{govindasamy2023nitrogen}. While our primary focus is on cover cropping, we further show that the proposed methods generalize well to other crops (e.g., wheat, corn) and management contexts. Previous remote sensing approaches to estimate cover crop biomass and N have relied on site-specific low-resolution satellite datasets \cite{deines2023recent, hively2020estimating, kc2021assessment}, dominated by the use of Sentinel-2 \cite{do2024estimating, fan2020winter, gao2020detecting, goffart2021field, thieme2020using, xia2021estimating} or UAS datasets \cite{holzhauser2022estimation, roth2018predicting, yuan2019unmanned, yuan2021advancing}, limiting precision and scalability of these approaches across space and time. In contrast, our neural network-based super-resolution framework integrates low-resolution satellite and high-resolution UAS imagery when available and performs robustly even when only one modality is present. We evaluated this framework on multiple treatment datasets and addressed the following key questions: (1) What type of UAS imagery (RGB, multispectral or hyperspectral) is most effective for estimating forage biomass yield and N content? (2) How does it compare with Sentinel-2 in terms of prediction accuracy? (3) Does improving the spatial resolution of Sentinel-2 and the spectral range of UAS RGB enhance the prediction of biomass and N? (4) Can the resulting models be extended across space and time, particularly to regions with limited UAS coverage or persistent cloud cover? These questions guide the experimental evaluation in the sections that follow and highlight the complementary role of UAS and satellite data in scalable, cost-effective precision management strategies. 

Section~\ref{sec:methods} introduces the datasets used in this study and details the proposed end-to-end super-resolution workflow. Section~\ref{sec:results} presents the results, including both the quality of super-resolved image reconstruction and its applications in precision farming. In Section~\ref{sec:discussion} we interpret the findings and discuss their broader implications. Section~\ref{sec:limitations} highlights a few limitations and suggests directions for future research. Finally, Section ~\ref{sec:conclusions} concludes the paper by summarizing the main contributions and highlighting the broader relevance of the proposed approach. 

\section{Materials and methods}
\label{sec:methods}
\subsection{Experimental design and site description}
\begin{figure}[h]
  \centering
  \includegraphics[width=1.0\textwidth]{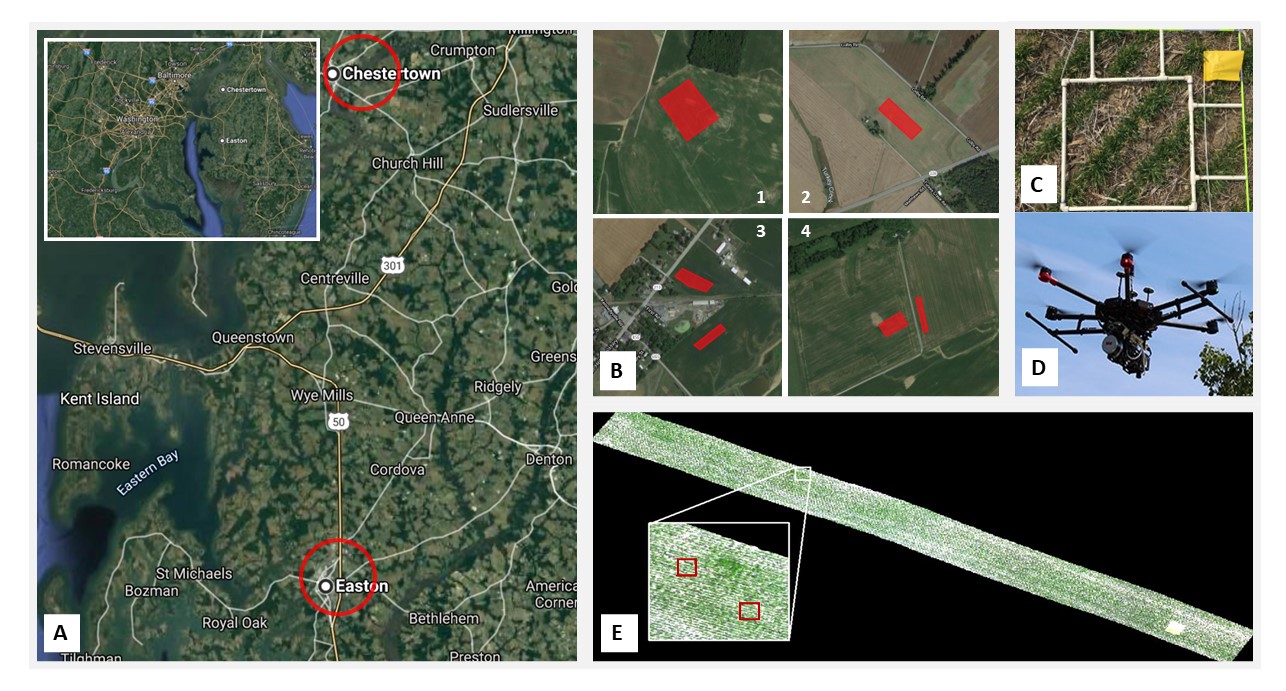}
  \caption{(A) Cover cropping sites - two farms are near Chestertown, MD and two farms near Easton, MD, studied over three time periods, from December 2018 to March and April 2019; (B) Images 1-4 are showing four sites: Fields A, B, D, E. 
  Each field in a site was divided for three cover crop treatments: fall seed drill, aerial seeding, and 3) none; (C) Cover crop characteristics measured: biomass yield, N content within a quadrat (0.5 m x 0.5 m); (D) UAS - DJI Matrice 600 Pro equipped with a Headwall Nano-Hyperspec [VNIR 400–1000 nm] and Velodyne VLP-16 LiDAR Puck LITE; Flight specs: <10 m/s, 40–50 m above ground level; (E) Hyperspectral data collected in one flight path. The locations of two quadrats are marked in red.}
  \label{fig:fig1}
\end{figure}

The study seeks to demonstrate the ability of the proposed super-resolution system in analyzing cropping practices in the upper [UCB] and lower Chesapeake Bay [LCB] region of the United States. The Eastern Shore of Maryland dominates row crop production in the LCB region. In the LCB we focused on cover crops within the dominant corn soybean crop rotation, and a diversity of soils and tillage practices common to the LCB. We included both standard and experimental cover cropping practices in the LCB. The LCB experiments were conducted in collaboration with four commercial grain farmers in Talbot and Kent counties. Farmers used their preferred crop rotation for each field (Figure~\ref{fig:fig1}). In each field, we imposed three cover crop management systems \cite{sedghi2023aerial}: 1) extended cover crop growing season by aerially interseeding the cover crop several weeks prior to cash crop harvest in fall and terminating it at cash crop planting in spring (Extended), 2) traditional cover crop growing season by drilling seed after cash crop harvest in fall and terminating it 3-4 weeks before cash crop planting in spring (Standard), and 3) no cover crop control (No cover). The cover crop was a three-species mixture with the farmer selecting a species from each group (brassica, legume, and cereal). In the UCB a dairy crop rotation is common and includes corn and harvested cover crops such as rye and triticale. We also included Miscanthus, switchgrass, and a mix of plant species used in the Conservation Reserved Program (CRP) \cite{adler2024modeling}.

\subsubsection{Prediction targets: cover crop biomass yield and N content} 
Details on the experimental design of cover crop fields, sample timing and location of fields, and cover crops biomass sample collection and processing for N measurements are provided in \cite{sedghi2023aerial}. We conducted hyperspectral flights over different fields in Maryland
(sites A and E near Easton Maryland and sites B, D and I near Chestertown Maryland) in December 2018, March and April 2019 using a Headwall Nano-Hyperspec [VNIR 400–1000 nm] (see details below). Each field was divided into three sections and each section was randomly assigned to flown, drilled, or no cover crop control treatments.  The field designations A, B, D, E, and I are the same as used in \cite{sedghi2023aerial}.

\subsection{The end-to-end framework: data processing, super-resolution, and application}

Our spatiotemporally scalable and an end-to-end image fusion system (see Figure~\ref{fig:proposed_system}) begins with spectral alignment of low-resolution satellite imagery -- specifically, simulating high-resolution Sentinel-2 data using hyperspectral UAS control data. This is followed by pixel-level alignment and image-to-image registration between the UAS and satellite imagery. Once aligned, we apply super-resolution techniques to enhance both spatial and spectral fidelity. The resulting high-resolution imagery are then applied to downstream agricultural tasks, such as biomass yield or nitrogen content estimation.  

\subsubsection{Hyperspectral control data}
\begin{figure}
  \centering
  \includegraphics[width=0.9\textwidth]{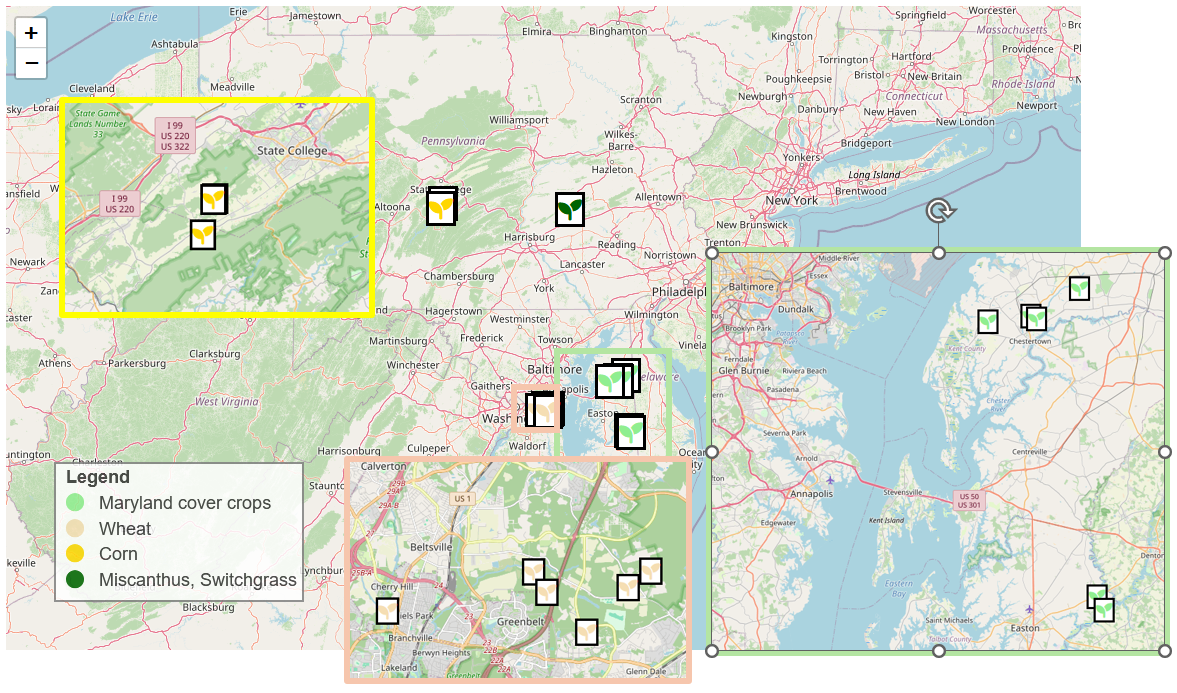}
  \caption{Hyperspectral flights locations.  The study sites are marked as "Maryland cover crops" in light green, while the corn, wheat and miscanthus/switchgrass crop images are marked in gold, beige and dark green respectively.}
  \label{fig:locations}
\end{figure}

To measure the generalization performance of the spectral extension models, we used a larger curated set of hyperspectral images collected from the same Headwall Nano-Hyperspec camera spread over a larger cross section of Maryland and Pennsylvania.  These images are from fields growing the crops corn, miscanthus, switchgrass, CRP and wheat in contrast to the three-species cover crop from the Maryland study sites.  The hyperspectral flight locations for the Maryland study sites and the images of corn, wheat and miscanthus are shown in Figure~\ref{fig:locations}.  These hyperspectral flights were spread over a period of time between 2018 and 2024.  We list the different sites used in this paper along with approximate location in Table~\ref{tab:field_list}.

\begin{table}[ht]
\caption{List of all locations where hyperspectral data was collected.  Sites A, B, D, E, and I were from the study sites and sites W:A-W:F where flown at Beltsville Agricultural Research Center in Beltsville Maryland.}
\label{tab:field_list}
\centering
\setlength\tabcolsep{2.0pt}
\begin{tabular}{llcl}
\toprule
 Nearest city & Date of flight(s)  & Crop & Site \\
\cmidrule(lr){2-4}
Easton, MD    & 5/12/18, 3/20/19, 4/16/19  & cover crop  & A, E\\
   & 11/21/19  &   & E\\
Chestertown, MD & 6/12/18, 3/20/19, 4/16/19 & & B, D, and I\\
 & 11/20/19 & & B, H\\
\cmidrule(lr){1-1}
Wariors Mark, PA & 7/29/20, 8/3/22  & corn & C:A\\
Pennsylvania Furnace, PA & 7/8/24 & & C:B\\
Leck Kill, PA & 6/8/20, 9/23/20 & miscanthus, switchgrass and CRP & M:A\\
Beltsville, MD & 4/15/20& wheat & W:A, W:B, W:C, W:D, W:E\\
 & 4/17/20&  & W:F\\
 \bottomrule
\end{tabular}
\bigskip
\end{table}

\subsubsection{Satellite image spectral alignment} 
\label{sec:spectral_alignment}
\begin{figure}[h!]
  \centering
  \includegraphics[width=0.8\textwidth]{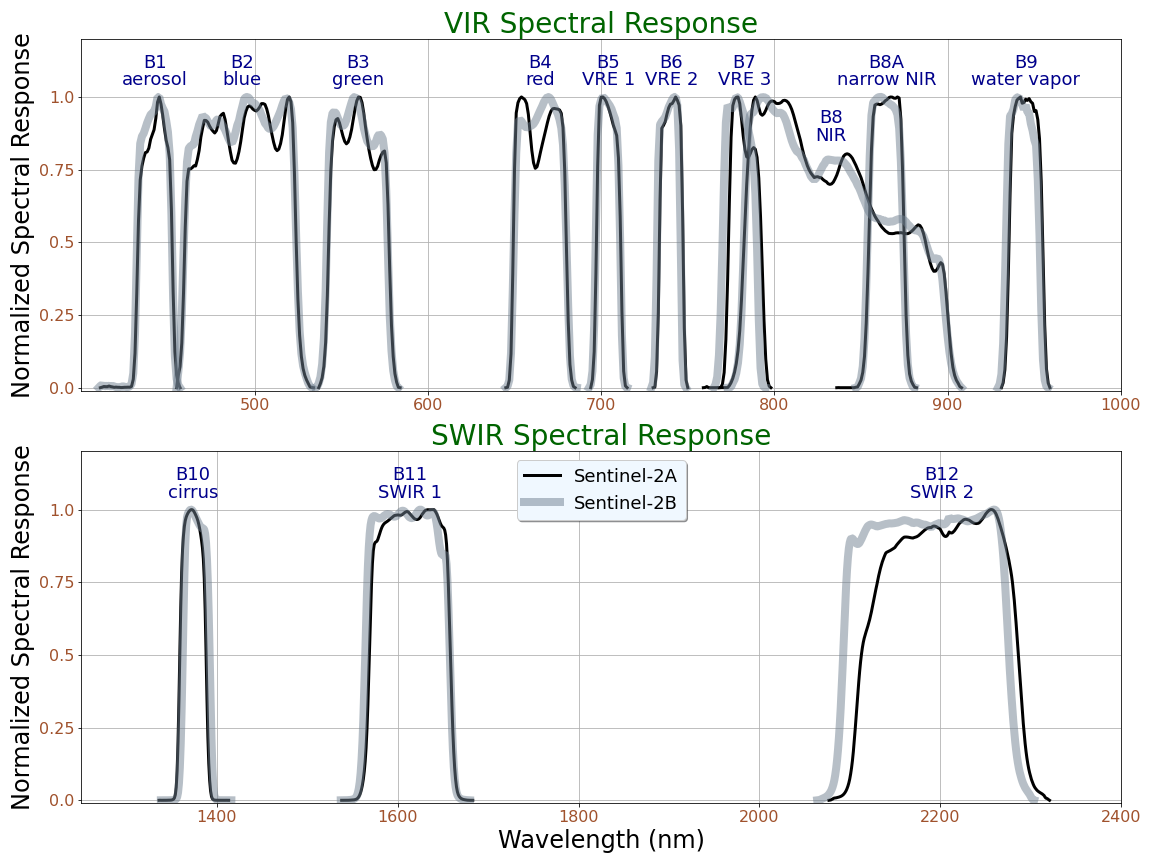}
  \caption{Measured normalized spectral response function (SRF) for Sentinel-2A and Sentinel-2B (S2-SRF) for VNIR bands used in this study.}
  \label{fig:fig2}
\end{figure}

\begin{figure}[h!]
  \centering
  \includegraphics[width=1.0\textwidth]{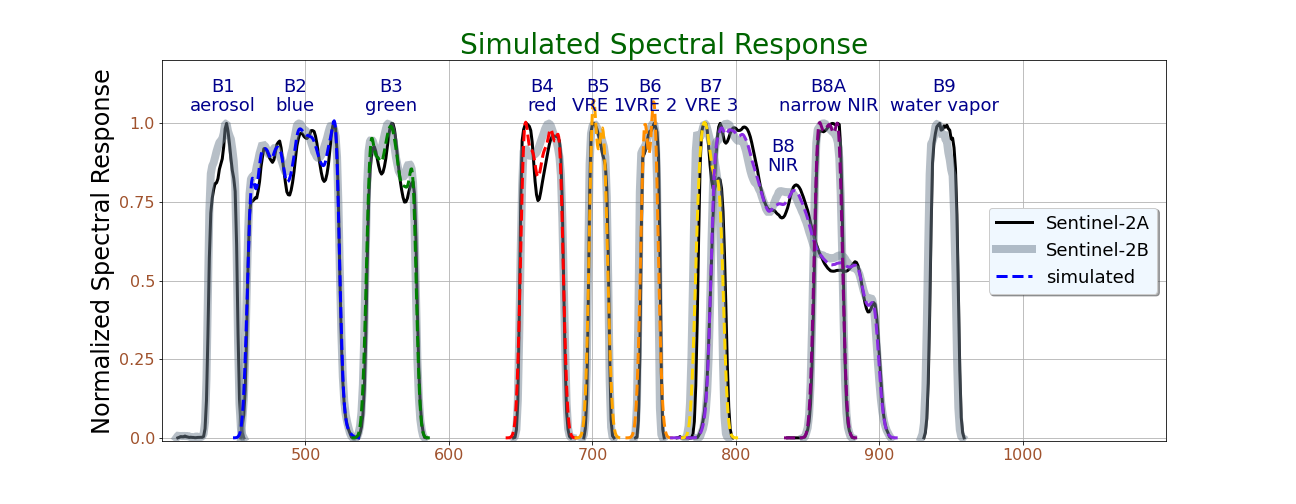}
  \caption{Simulated Spectral response average of 8 Sentinel-2 VNIR bands using the Headwall Nano-Hyperspectral camera and the published S2-SRF. The dashed colored lines correspond to the non-negative linear regression estimate of the spectral response function.}
  \label{fig:fig3}
\end{figure}

To train a super-resolution model for Sentinel-2 we need high-resolution ground-truth data. Sentinel-2 carries a pushbroom multispectral instrument (MSI), which measures the reflected radiance in thirteen spectral bands. We have access to neither the MSI nor to higher-resolution images taken with the MSI. However, the spectral bands are characterized by the measured spectral response function found in \cite{COPE-GSEG-EOPG-TN-15-0007} and shown in Figure~\ref{fig:fig2}. Given data from a hyperspectral sensor, we took a weighted average of its narrow spectral bands to simulate the spectral response of the MSI. However, the Sentinel-2 MSI will be affected by greater atmospheric noise compared to a UAS. The Headwall Nano-Hyperspec [VNIR 400–1000 nm] was used to collect hyperspectral UAS images. The Headwall Nano-Hyperspec provided 269 bands in the VNIR with a bandwidth given by a FWHM (Full Width at Half Maximum) of 6 nm. Furthermore, Headwall kindly provided us with the relative spectral response as a function of the wavelength. The much narrower bandwidth of the hyperspectral data means that we can approximate the MSI spectral response function by a weighted average of the hyperspectral bands. Although linear regression is a possible approach, we noted that the spectral response is nonnegative, so we resorted instead to the nonnegative linear squares (NNLS) problem to prevent negative values.  The NNLS minimized the reconstruction error for a linear combination of hyperspectral bands against the published values of the measured spectral response function (S2-SRF). We used the FORTRAN code published in the book \cite{lawson1995solving} that solves the Karush-Kuhn-Tucker conditions for the non-negative linear squares problem.  Specifically, we access the code through the Scipy Optimize library \cite{https://doi.org/10.5281/zenodo.6940349}. The spectral response for the 8 Sentinel-2 bands B2-B8 and B8A in the VNIR range was approximated and the results can be seen in Figure~\ref{fig:fig3}. Among the 269 hyperspectral bands, only 126 were activated with nonzero weights for the 8 simulated Sentinel-2 bands. Note that the remaining bands are either in the short-wave infrared (SWIR) range not covered by the hyperspectral camera or are targeting atmospheric observations at a resolution of 60 m.

\subsubsection{UAS pixel alignment}
\begin{figure}[h!]
  \centering
  \includegraphics[width=0.8\textwidth]{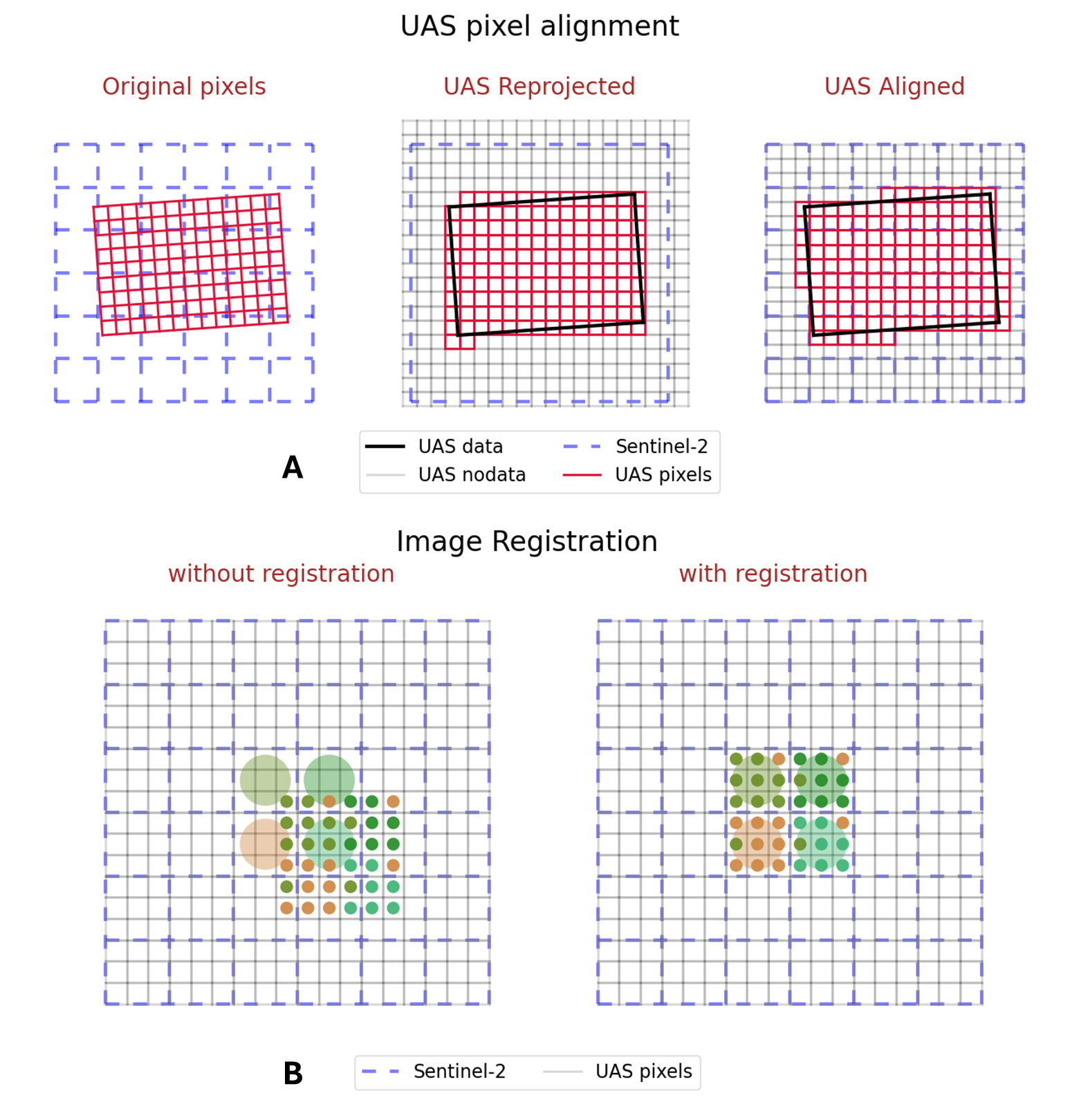}
  \caption{(A) Transforming and aligning the UAS image to the Sentinel-2 image.  The middle image shows that simply changing the coordinate reference system (reprojecting) may leave the Sentinel-2 pixel corners in the middle of a UAS pixel.  A second transformation shifts the origin so that the upper left corners of the images align. 
  (B) An illustration of the image registration process.  The small circles represents UAS pixels, and the large circles Sentinel-2 pixels.  Linear regression is used to determine the shift for the co-registration of the images allowing us to move from the image on the left to that of the right.}
  \label{fig:uas-align}
\end{figure}

The spectral alignment process provides a higher spatial resolution image based on the hyperspectral UAS image corresponding to the Sentinel-2 MSI sensor, but does so in the geometry of the original UAS.  For the purpose of training neural network models, we also need to align the pixels of the UAS image with the Sentinel-2 image.  We do so in two stages; (1) transform the UAS image to the Sentinel-2 reference coordinate system, and (2) align the upper left corner of the UAS image to that of the upper left corner of a Sentinel-2 pixel.  The second step ensures that no UAS pixel straddles two or more Sentinel-2 pixels and also that every Sentinel-2 pixel is fully covered by UAS pixels.  In this process, it is important to leave the more vulnerable low-resolution image completely unchanged; therefore, the transformations are only applied to the UAS image.  Figure~\ref{fig:uas-align}A illustrates the process.  The figure's illustrations are meant as a schematic and show roughly three by three pixels per Sentinel-2 pixel.  In our application, the UAS pixel resolution is 0.125 m versus 10 m for the Sentinel-2 pixels, giving $80\times 80$ UAS pixels for each Sentinel-2 pixel.

\subsubsection{Image registration}
If the Sentinel-2 and UAS pixels are perfectly aligned with their respective reference coordinate systems, the pixel alignment process should suffice to align the images.  However, it is unreasonable to expect the Sentinel-2 pixels to have a location accuracy lower than its pixel resolution.  Moreover, the UAS pixel location accuracy depends on the quality of the equipment used, and we need to provide a working solution for any reasonable UAS.  To correct errors from the pixel alignment process, we resort to a classic image registration methodology.  However, there are two obstacles standing in our way.  Firstly, the spatial resolutions of the two images are different, with the Sentinel-2 image being much coarser than the UAS image.  Secondly, the images have different bands that correspond to different spectral response curves.  The low resolution of the Sentinel-2 pixels precludes the use of key-points to co-register the images.  Instead, we consider all translations of the UAS image by any amount within $\pm1$ Sentinel-2 pixel and utilize linear regression to measure the difference between the UAS and Sentinel-2 image.  The shift with the lowest regression error determined the final image registration.  Figure~\ref{fig:uas-align}B illustrates the process.  The small circles represent some of the UAS circles, while the large circles represent Sentinel-2 pixels.  We see how the process can lead to a better correspondence between the images.  

\begin{figure}
\centering
  \includegraphics[align=c, width=0.55\textwidth]{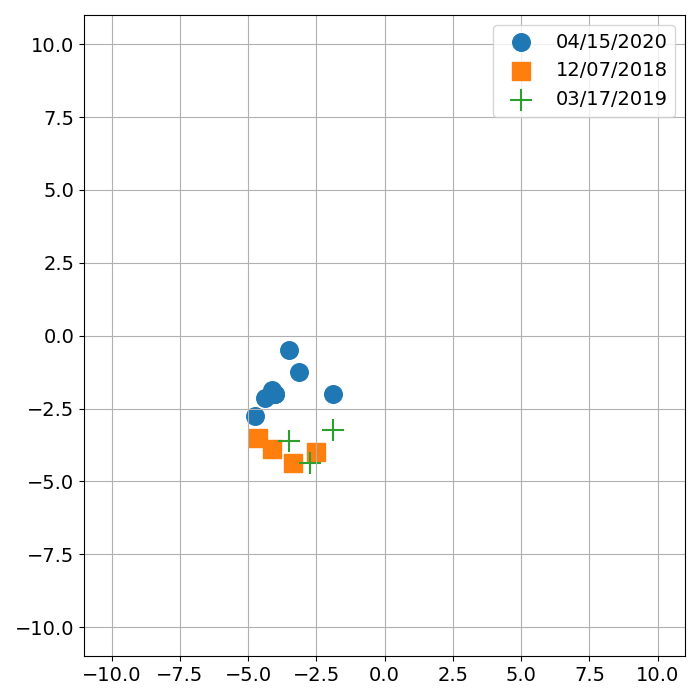}
  \caption{Optimal shift values found for each image-to-image registration.  Sentinel-2 tiles cover more than 100 km $\times$ 100 km and images with the same marker type correspond to the same Sentinel-2 tile.}
  \label{fig:registration-shifts}
\end{figure}

Figure~\ref{fig:registration-shifts} shows the amount of image translation for a collection of images as determined by the image-to-image registration process.  Each marker type and color correspond to the same larger Sentinel-2 image (also known as a \textit{granule} or {\em tile}).  Given the high accuracy of the location of our hyperspectral images, it is reasonable to assume that images corresponding to the same Sentinel-2 tiles should be shifted by the same amount, and indeed some clustering of the calculated shift values agrees with that assumption.  Under the optimistic assumption that the unknown real optimal shift corresponds to the median of the clusters we can estimate the registration shift error to be approximately 0.81 m compared to the no-image registration error which is approximately 4.6 m. 

\begin{figure}
\centering
  \includegraphics[align=c, width=0.95\textwidth]{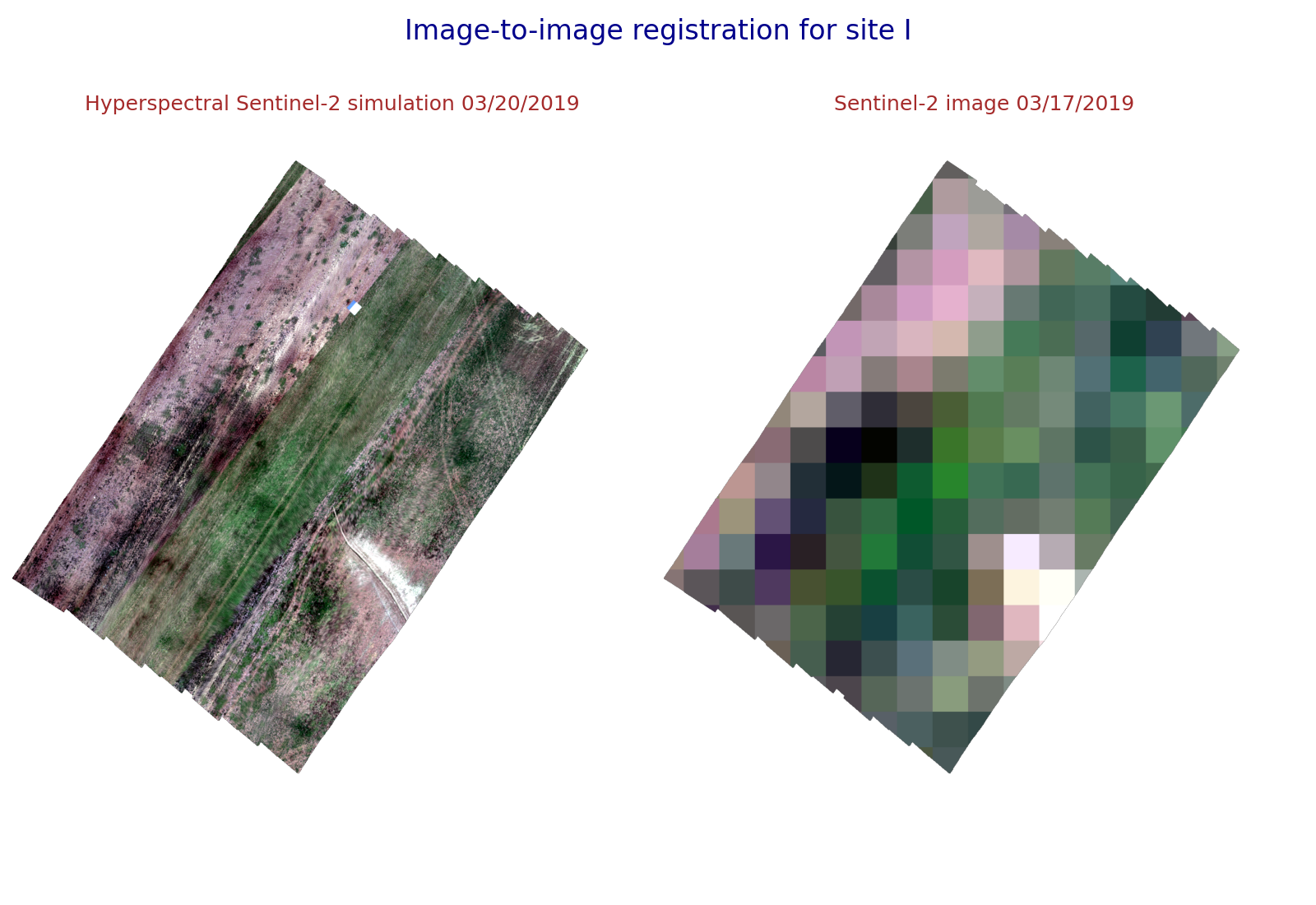}
  \caption{An example image-to-image registration for site I.  The invalid (no-data) hyperspectral pixels are also masked out in the Sentinel-2 image to aide comparison.}
  \label{fig:example-registration}
\end{figure}

Figure~\ref{fig:example-registration} shows the image-to-image registration for site I.  Note how well the colors align and how larger features with distinct colors are reflected in the Sentinel-2 pixels.  This example also illustrates how much more detailed the hyperspectral-derived Sentinel-2 simulation is compared to the Sentinel-2 image itself.   

\subsection{Super-resolution system for fusion and precision farming application}
\begin{figure}
  \centering
  \includegraphics[width=0.9\textwidth]{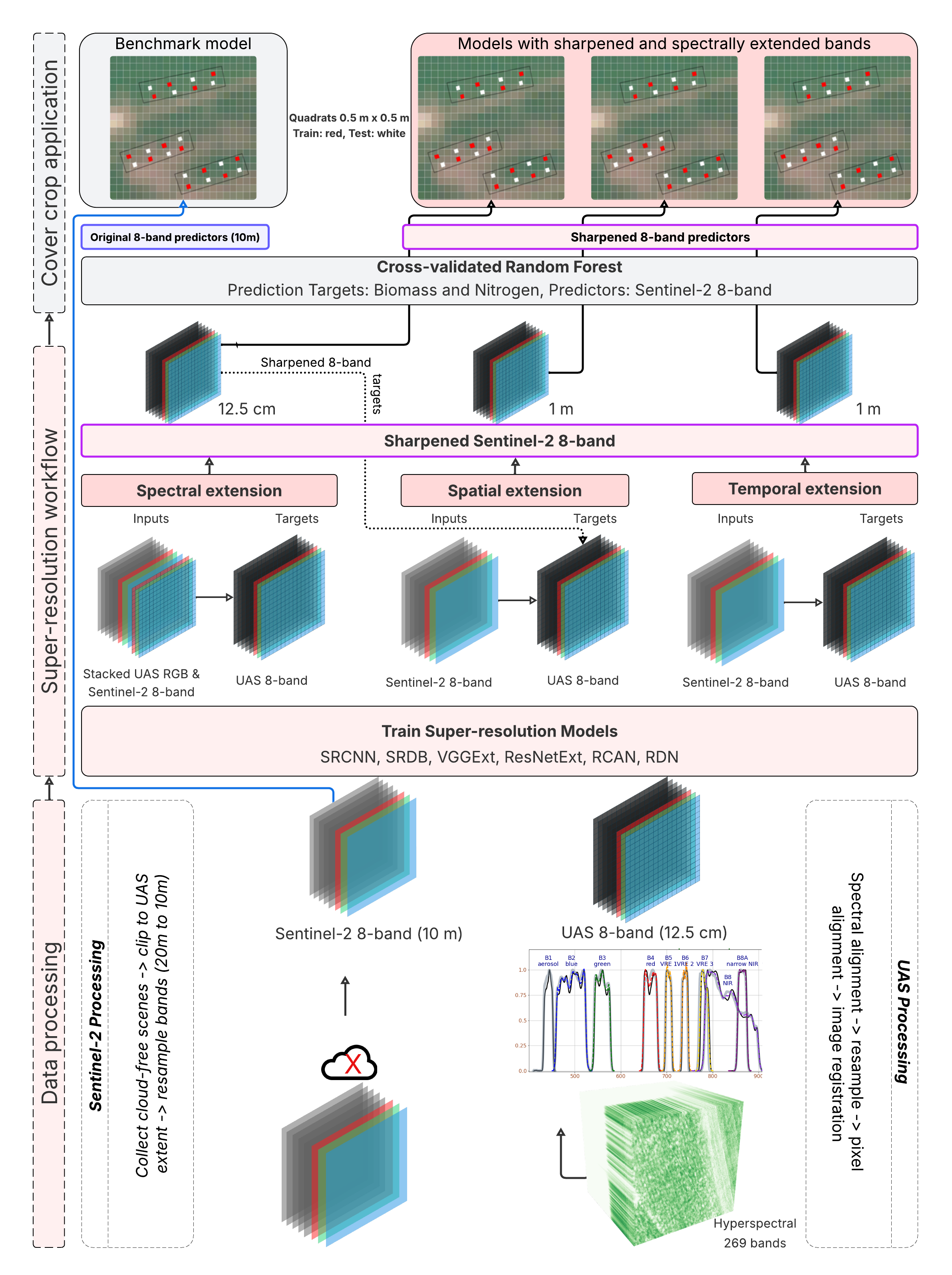}
  \caption{The proposed end-to-end system for data preparation, super-resolution fusion and predictive modeling workflow to assess complementarity of multi-source imagery in precision agriculture.}
  \label{fig:proposed_system}
\end{figure}

We used super-resolution image fusion to reconstruct high-resolution Sentinel-2 bands in the VNIR range and to extend UAS RGB imagery to include VRE and NIR spectral regions. Figure~\ref{fig:super-resolution-frameworks} shows a comparison of common super-resolution modeling frameworks and the spectral SRCNN introduced in this paper. We then evaluated whether the reconstructed sub-meter and 1 m resolution multispectral (MS) imagery improved biomass and N prediction compared to 10m Sentinel-2 and sub-meter UAS RGB data. Figure~\ref{fig:proposed_system} illustrates the overall super-resolution system and the application modeling framework.

As introduced in section~\ref{sec:extensions}, we developed three distinct super-resolution scenarios -- spectral, spatial, and temporal extensions -- considering their real-world applications in cost-effective precision farming. These scenarios account for multispectral data variability caused by plant phenology and field-specific spatial factors such as topography and soil type. For example, in the \textbf{spectral extension} scenario, where only RGB UAS imagery is available, we fuse it with 10m/20m Sentinel-2 VNIR bands to produce sub-meter resolution 8-band MS data. This extends RGB into the VRE and NIR spectral domains. In the \textbf{spatial extension} case, we generate high-resolution MS imagery for farm areas lacking UAS data by using outputs from the spectral extension model, eliminating the need to survey the entire field. In the \textbf{temporal extension} scenario, we generate high-resolution imagery for the same field at a different time when no recent UAS data are available. Table~\ref{tab:data_splits} details the data splits used for each scenario, capturing spatial and temporal variability across different fields and time periods.  

To train and evaluate these super-resolution models: 
\begin{itemize}
    \item Spectral extension model used all available original Sentinel-2 VNIR, UAS weighted VNIR, and UAS weighted RGB-only (see Section \ref{sec:spectral_alignment}) data from all fields and temporal periods across MD and PA to capture spatial and temporal variability owing to plant phenology. Site-level cross-validation ensured that test data were excluded during training and validation.
    
    \item  Spatial extension model used a subset of available data from different fields for model training and validation. Model inference was conducted on a completely separate set of fields.   
    \item  Temporal extension model was trained on field data from one season and tested on the same fields from a different season. 
\end{itemize}

\begin{table}[h!]
\centering
\caption{Super-resolution modeling data splits by extension type, site, geography, and timeline.}
\label{tab:data_splits}
\begin{tabular}{lccc|ccc}
\toprule
\multirow{2}{*}{\textbf{Extension}} & \multicolumn{3}{c|}{\textbf{Train \& Validation Data}} & \multicolumn{3}{c}{\textbf{Test Data}} \\
\cmidrule(lr){2-4} \cmidrule(lr){5-7}
 & \textbf{Sites} & \textbf{Geography} & \textbf{Timeline} & \textbf{Sites} & \textbf{Geography} & \textbf{Timeline} \\
\midrule
Spectral & A, B, I, D, E & MD & Mar--Apr 2019 & A, B, I, D, E & MD & Mar--Apr 2019 \\
 &  &  & & \makecell[l]{W:A, W:B, W:C,\\ W:D, W:E, W:F,\\ C:A, C:B, M:A} & MD \& PA & See Table \ref{tab:field_list} \\
Spatial  & All sites except A, E        & MD & See Table \ref{tab:field_list} & A, E         & MD         & Mar--Apr 2019 \\
Temporal & A, D, E        & MD & Mar--Nov 2019 & A, D, E      & MD         & Dec 2018 \\
\bottomrule
\end{tabular}
\end{table}

\subsubsection{Super-resolution methods}
\label{sec:super-resolution-methods}
For the spatial and temporal extension tasks, we utilized techniques from the super-resolution literature. However, for the spectral extension task we could only use super-resolution when we were confined to the VNIR spectrum that we could simulate with our hyperspectral camera. 

\begin{figure}[h!]
  \centering
  \includegraphics[align=c, width=0.7\textwidth]{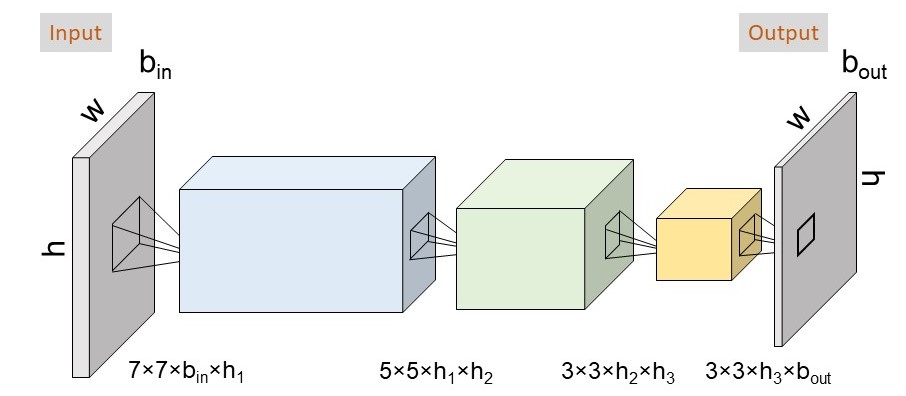}
  \caption{The SRCNN network architecture.}
  \label{fig:SRCNN}
\end{figure}

\textbf{Super-resolution Convolutional Neural Networks (SRCNN):} 
SRCNN \cite{dong2014learning} was one of the pioneering methods that applied deep neural networks to single-image super-resolution (SISR). It demonstrated that end-to-end learning is effective in sharpening blurred photographs. Since SRCNN many other more effective methods have been developed such as residual dense networks (RDN) \cite{zhang2018residual}, residual channel attention network (RCAN) \cite{zhang2018image}, super-resolution through repeated refinement (SR3) \cite{saharia2022image}. But here we focus on how super-resolution can be applied to precision farming and not on which super-resolution method is the best. Figure~\ref{fig:SRCNN} shows the specific SRCNN architecture we used for the spectral extension model in the results section. The SRCNN model consists of 3 components; 1) a feature extractor corresponding to the first layer that convolves the input image with a large kernel (7x7 or larger), 2) non-linear block(s) that applies non-linear activations to capture complex relationships in the feature maps, and finally a 3) reconstruction layer that transforms the feature maps back to the desired image shape. 

A detail regarding the SRCNN model is that it takes an up-sampled low-resolution image as input. This enables us to simply stack the upsampled Sentinel-2 images with high-resolution UAS RGB images for the spectral SRCNN model. For more modern super-resolution models, actual architecture changes are required to handle dual image input.  For our spatial and temporal extension models, we also used SRCNN, but with different kernel sizes.

\textbf{Other Spectral Extension Super-resolution Networks (VGGExt, ResNetExt, and SRDB):}
We have deviated from the original SRCNN architecture by not using batch normalization, and instead of the \texttt{ReLU} activation we use the \texttt{LeakyReLU} activation.  In general, we find that batch normalization hurts somewhat in the context of satellite images despite being widely used for computer vision for photographs.  We also found that larger super-resolution models simply failed on our task.  This is almost certainly due to the size of the training data.  Despite having collected 17TB of hyperspectral data, the information is greatly compressed in the data preparation process.  The overlap of flight strips almost duplicates the data and the simulation of the Sentinel-2 MSI sensor brings the number of bands from 269 to 8.  We also resample the 3 cm hyperspectral data to 12.5 cm data.  This process reduces the size of the training data by 3 orders of magnitude, which means that we cannot train very large networks and expect good performance.  We will see later that a state-of-the-art network like RCAN does not perform as well as SRCNN on our task. 
We are reaching out to other researchers to get access to more hyperspectral data, but ultimately one should expect to have to train models on small datasets as utilizing petabytes of hyperspectral data is highly impractical.  However, having a 17TB database of hyperspectral images does allow us to generate ground-truth data for other satellites (SuperDove, Landsat, Airbus, etc.) and target resolutions without having to collect new data.

For the spectral extension super-resolution task in which SRCNN operates directly on images in the target resolution (Sentinel-2 images are upsampled) we consider other networks such as VGG \cite{vgg}, ResNet \cite{resnet}, and DenseNet \cite{densenet}.  These networks can be deeper than SRCNN through the use of max-pooling or residual connections.  However, max-pooling has the drawback of leading to blurry images, so for the VGG architecture, we created the miniature model that we call VGGExt in Figure~\ref{fig:VGGExt}, where we only used one max-pooling layer and later a conjugate convolution for up-sampling.  The resulting network is much smaller than even the VGG11 model and has been modified to do spectral extension instead of image classification.  Similarly Figure~\ref{fig:ResNet} shows our ResNet inspired network where all batch normalization has been removed, the network has been reduced to essentially one residual block, and up-sampling components have been inserted at strategic locations to avoid loss of resolution in the inferred image.  For DenseNet, there is already a super-resolution network, RDN, based on its architecture.  However, we found RDN to be too large and we created a new small network that we call SRDB that encompasses features from DenseNet and uses kernel sizes inspired by our SRCNN model.  The interested reader can see the full model in Figure~\ref{fig:SRDB}. 

\subsubsection{Apply reconstructed imagery to predict cover crop biomass yield and quality}
We developed two sets of cross-validated random forest (RF)-based regression models to predict cover crop biomass yield and quality (assessed by N content) across different fields in the upper and lower Chesapeake Bay regions. The first set of RF models focused on exploring whether and how much the spatial and spectral resolution and the spectral range of the original MS (Sentinel-2 and UAS) and UAS hyperspectral imagery impact the predictive model performance in precision management practices. Insights from these experiments were then used to develop super-resolution models to reconstruct MS imagery with the desired spatial resolution and spectral range. The second set of RF models used these reconstructed high-resolution and broader spectral range data to predict biomass and N for the same fields. 

In the spectral extension experiment, the RF model used reconstructed spectral data from all eight datasets (i.e., two from site A, one from site B, two from site D, two from site E, and one from site I flights) that were generated by an eight-fold SRCNN model; each SRCNN model was trained on samples from seven datasets to infer on the remaining dataset. In that way, the ground-truth images corresponding to the reconstructed data did not participate in the SRCNN model training. This careful experimental protocol thus avoids data leakage between SRCNN training and testing, thus ensuring the fidelity of the reconstructed bands for the test site. The spatial and temporal extension SRCNN model training and testing sites were different, either spatially or temporally, thus no cross-validation approach was needed during super-resolution process. Table~\ref{tab:data_splits} Test Data list for MD sites shows which reconstructed Sentinel-2 imagery were used to extract 0.5 m x 0.5 m quadrat-level 8-band predictors for our RF models. A variable number of quadrats were used as data samples, ranging from 72 to 96, for the spectral, spatial, and temporal extension RF models.    

All RF regression models for spectral, spatial, and temporal extension scenarios were trained with 5-fold cross-validations to maximize the size of the training data. Model testing was performed on a randomly selected fold out of five folds. The root mean squared error (RMSE) and the R-squared statistical measures were calculated to compare the performance of the RF models. 

\section{Results}
\label{sec:results}
\subsection{Prediction of biomass yield and quality using original Sentinel-2 and UAS imagery}\label{sec:spectral_performance}

Table \ref{tab:org_models} presents cover crop biomass yield and quality estimation results from Random Forest (RF) models using Sentinel-2 and UAS imagery across varying spatial resolutions, spectral resolutions, and spectral ranges (\modelref{S2:RGB}-\modelref{hyp:S2:5m}). Overall, estimation accuracy improved with broader spectral coverage for both platforms. Sentinel-2-based RF models showed significant gains when the full VNIR range (8 bands) was used for biomass, and when SWIR bands were added (10 bands total) for N estimation (\modelref{S2:RGB}-\modelref{S2:10bands}). In contrast, UAS-based models improved up to the RGB range for biomass, and with five bands (RGB and three VRE) for N (\modelref{MX:RGB}-\modelref{MX:5bands}). With weighted 8-band data from Headwall hyperspectral 269 bands (based on the Sentinel-2 MSI spectral response function, as discussed in Section~\ref{sec:spectral_alignment}), we observed the same patterns of impacts of spectral range as the original Sentinel-2, however, RMSEs were much lower with the hyperspectral simulated Sentinel-2 8-band data (\modelref{hyp:S2:RGB}-\modelref{hyp:S2:8bands}). To further assess the effect of specific spectral ranges on model performances, hyperspectral 269-band data at varied wavelength ranges were evaluated in models \modelref{Range:50}-\modelref{Range:269}. Biomass yield estimation error improved but remained unaffected by the change in spectral ranges (397.9-1002.9 nm). For N estimation, however, broader spectral coverage remained beneficial, with the lowest RMSE (16.09 kg/ha) achieved using model \modelref{Range:269}, which included spectral information from bands spanning 397.9–1002.9 nm. 

The higher spectral resolution ($\sim$13 nm to $\sim$86 nm) of the 269-band hyperspectral data contributed to improved RF model performance for both biomass and N estimation (\modelref{Resolution:13nm} - \modelref{Resolution:86nm}). Model errors increased sharply with decreased spectral resolutions ($\sim$150 nm and $\sim$600 nm) (\modelref{Resolution:150nm} - \modelref{Resolution:600nm}). In terms of spatial resolution, \modelref{hyp:S2:0.125m} using simulated 8-band (0.125m) data produced the best performance for both targets. However, \modelref{hyp:S2:1m}, which used 1m resolution data, was not far behind, performed comparably well, suggesting that both sub-meter and 1 m resolutions are viable options for cover crop mapping. Notably, although \modelref{hyp:S2:5m} produced the highest RMSEs among these three models, it still outperformed the Sentinel-2 10 m band models (\modelref{S2:RGB}-\modelref{S2:10bands}) by a considerable margin.
 
\begin{figure}[ht]
  \centering
  \includegraphics[width=1.0\textwidth]{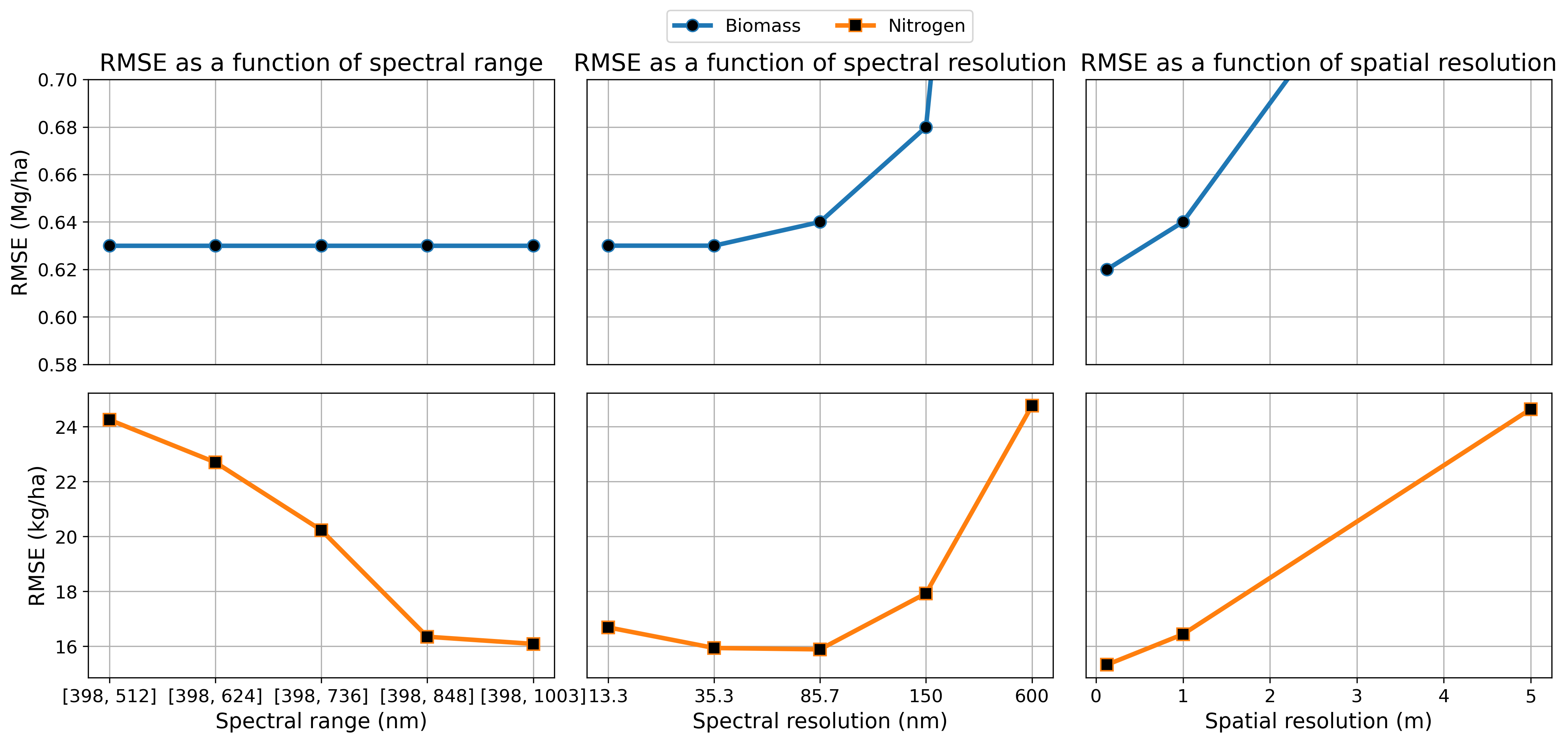}
  \caption{RMSEs of biomass (top row) and N estimations (bottom row) as functions of spectral range (M7-M11), spectral resolution (M12-M18) and spatial resolution (M22-M24).}
  \label{fig:fig8}
\end{figure}

Key insights from these experiments are as follows:

\begin{itemize}
    \item Spectral range is more important for estimating N content than for biomass yield. 
    \item Higher spectral resolution improves both biomass and N estimations (Figure~\ref{fig:fig8}). It appears that by improving the spectral resolution of the RGB and VRE-NIR bands, we can capture important information about vegetation health.
    \item Reducing spatial resolution from meters to decimeters (e.g., from 1m to 0.125m) yields substantial gains, potentially influenced by the scale of our experimental design and quadrat size. 
    \item For RedEdge-MX, expanding from RGB to 5-band yields smaller gains compared to Sentinel-2, but when simulating Sentinel-2 bands using hyperspectral data, the performance improvement is comparable to actual Sentinel-2. 

\end{itemize}

Given the cost and impracticality of flying large areas with UAS and deploying hyperspectral sensors, we next evaluated super-resolution based reconstructed high-resolution imagery as a cost-effective alternative. These reconstructions outperformed raw Sentinel-2 and UAS RGB data in downstream prediction tasks, offering a practical solution for scalable precision agriculture. 

\begin{table}[h!]
\caption{Impacts of spatial and spectral resolution and spectral range (25 bands per range) on model performance. The native spatial resolution of Sentinel-2A is 10 m for RGB+NIR and 20 m for VRE and Narrow NIR bands, while the spatial resolution for Micasense RedEdge MX and Hyperspectral images is $\sim$ 3 cm.}
\setlength\tabcolsep{2.0pt}
\centering
\begin{tabular}{llcccc}
\toprule
\textbf{Tag} & \textbf{Sensor} & \multicolumn{2}{l}{\textbf{Biomass}} & \multicolumn{2}{l}{\textbf{Nitrogen}} \\
 &   & $R^2$      & RMSE (Mg/ha) & $R^2$       & RMSE (kg/ha) \\
\cmidrule(lr){3-4}\cmidrule(lr){5-6}
\modellabel{S2:RGB} 
& Sentinel-2A RGB &  18.5    & 1.33        & -1.4     & 36.79       \\
\modellabel{S2:5bands} 
& Sentinel-2A 5-band  & 51.7    & 1.04        & 37.5     & 28.81      \\
\modellabel{S2:8bands} 
& Sentinel-2A 8-band   & 69.1    & 0.87        & 61.1     & 23.88      \\
\modellabel{S2:10bands} 
& Sentinel-2A 8-band \& SWIRs & 67.9    & 0.88        & 64.1     & 23.09      \\
\modellabel{MX:RGB} 
& RedEdge-MX RGB  & 84.2    & 0.65        & 71.1     & 21.14        \\
\modellabel{MX:5bands} 
& RedEdge-MX 5-band  & 83.9    & 0.66        & 76.2     & 19.79      \\
& & & & & \\
& \textbf{Hyperspectral Sentinel-2 simulations}  [397.9, 1002.9] (nm) & & & & \\
\cmidrule(lr){2-2}
\modellabel{hyp:S2:RGB} 
& 3 bands: RGB & 80.7    & 0.72        & 59.9     & 25.10\\
\modellabel{hyp:S2:5bands} 
& 5 bands: RGB + VRE + NIR  & 84.2    & 0.65        & 77       & 19.19\\
\modellabel{hyp:S2:8bands} 
& 8 bands: RGB + 3x VRE + NIR + Narrow NIR & 85.2    & 0.64        & 89.6     & 14.93 \\
& & & & & \\
& \textbf{Spectral Range (25 bands)} & & & & \\
\cmidrule(lr){2-2}
\modellabel{Range:50} 
& Wavelength: [397.9, 511.8] (nm)  & 85.4    & 0.63         & 60.8     & 24.25   \\
\modellabel{Range:100} 
&  [397.9, 623.9] & 85.4    & 0.63        & 65.6     & 22.70  \\
\modellabel{Range:150} 
&  [397.9, 736] & 85.7      & 0.63        & 74.6     & 20.23 \\
\modellabel{Range:200} 
&  [397.9, 848.2] & 85.8      & 0.63        & 86.3     & 16.35 \\
\modellabel{Range:269} 
&  [397.9, 1002.9] & 85.8    & 0.63        & 87.4     & 16.09 \\
& & & & & \\
& \textbf{Spectral Resolution}  [397.9, 1002.9] (nm) & & & & \\
\cmidrule(lr){2-2}
\modellabel{Resolution:13nm} 
& resolution $\sim$ 13nm & 85.7    & 0.63         & 86.2     & 16.69\\
\modellabel{Resolution:35nm} 
& resolution $\sim$ 35nm & 85.7    & 0.63         & 87.4     & 15.94\\
\modellabel{Resolution:86nm} 
& resolution $\sim$ 86nm & 85.7    & 0.64        & 88.4     & 15.89\\
\modellabel{Resolution:150nm} 
& resolution $\sim$ 150nm & 85.5    & 0.68        & 87.7     & 17.93\\
\modellabel{Resolution:600nm} 
& resolution $\sim$ 600nm & 44.7    & 1.12        & 58.6     & 24.76\\
& & & & & \\
& \textbf{Spatial Resolution}  (simulated Sentinel-2 8-band) & & & & \\
\cmidrule(lr){2-2}
\modellabel{hyp:S2:0.125m} 
& 0.125m & 85.6    & 0.62        & 89.2     & 15.33 \\
\modellabel{hyp:S2:1m} 
& 1m & 84.8    & 0.64        & 85.8     & 16.44 \\
\modellabel{hyp:S2:5m} 
& 5m & 72.5    & 0.84         & 59.6     & 24.64 \\
\bottomrule
\end{tabular}
\label{tab:org_models} 
\end{table}

\subsection{SRCNN reconstructed MS data for spectral, spatial, and temporal extensions}

In the spectral extension scenario, the RGB bands from the UAS are used alongside Sentinel-2's bands to generate high-resolution (0.125 m) MS images. Our proposed spectral SRCNN can produce accurate, high-resolution (0.125 m) images containing 8 Sentinel-2 spectral bands across the VNIR range with exceptional fidelity. This approach can be useful in scenarios where the farmer has access to an RGB UAS but not to an MS or hyperspectral UAS. In contrast, the spatial extension scenario enables the farmer to not fly the UAS over all fields, and instead use Sentinel-2 imagery together with a pretrained super-resolution/SRCNN model (trained on data from different fields). Similarly, the same approach can be used in the temporal extension scenario to reconstruct high-resolution Sentinel-2 data from a different time period. In addition to cost savings, spatial and temporal extension scenarios allow the farmer to balance the time of flying against other crop management activities, which is particularly useful for the most labor intensive periods.

The spectral fidelity of the reconstructed high-resolution Sentinel-2 data is assessed using different matrices including mean squared error (MSE), root-mean-square error (RMSE), mean absolute error (MAE), and peak signal-to-noise ratio (PSNR). PSNR is a good measure for how well the color is retained (spectral fidelity). Table \ref{tab:tab2} shows that the Spectral-SRCNN resulted in reconstructed images with higher spatial and spectral fidelity compared to our other extension models. PSNR values below 24 are considered low, between 24 and 30 are moderate, and above 30 correspond to only minor differences between the reconstructed image and the ground-truth (in contrast, nearly loss-less compression system has a PSNR around 40). We postulate from this result that the Spectral-SRCNN approach is more likely to translate to other applications in precision management. 

\begin{table}[h!]
\centering
\caption{Accuracy assessment of SRCNN based image reconstruction for spectral, spatial and temporal extensions. The month column represents the UAS flight dates.}
\begin{tabular}{llllll}
\toprule
Extension Scenario & Site & Month  & RMSE  & MAE   & PSNR \\
\cmidrule(lr){1-1} \cmidrule(lr){2-6}
Spectral           & \texttt{A} & 3/20/19 & 0.0164 & 0.0077 & 35.69\\
                   &            & 4/16/19 & 0.0149 & 0.0069 & 36.56\\
                   & \texttt{B} & 3/20/19 & 0.0415 & 0.0188 & 27.64\\
                   & \texttt{I} &         & 0.0233 & 0.0105 & 32.67\\
                   & \texttt{D} &         & 0.0234 & 0.0110 & 32.61\\
                   &            & 4/16/19 & 0.0352 & 0.0170 & 29.08\\
                   & \texttt{E} & 3/20/19 & 0.0242 & 0.0116 & 32.33\\
                   &            & 4/16/19 & 0.0229 & 0.0101 & 32.82\\
                   & Average    &          & \textbf{0.0252} & \textbf{0.0117} & \textbf{32.42}\\
\cmidrule(lr){2-6}
Spatial            & \texttt{A}   & 3/20/19 & 0.0274 & 0.0175 & 31.25 \\
                   &              & 4/16/19 & 0.0313 & 0.0188 & 30.09 \\
                   & \texttt{E}   & 3/20/19 & 0.0371 & 0.0232 & 28.61 \\
                   &              & 4/16/19 & 0.0406 & 0.0229 & 27.83 \\
                   & Average      &         & \textbf{0.0341} & \textbf{0.0206} & \textbf{29.44} \\
\cmidrule(lr){2-6}
Temporal           & A     & 12/05/18 & 0.0545 & 0.0283 & 25.28 \\
                   & D     & 12/06/18 & 0.0359 & 0.0209 & 28.90 \\
                   & E     & 12/05/18 & 0.0479 & 0.0270 & 26.40 \\

                   & Average &        & \textbf{0.0461}  & \textbf{0.0254} & \textbf{26.86}\\
\bottomrule
\end{tabular}
\label{tab:tab2}
\end{table}





\subsection{Comparison of different super-resolution models}
We made the SRCNN network the core model for high-resolution MS image reconstruction. SRCNN was the first neural network-based super-resolution model and has long been surpassed by larger, deeper, more complex, and more accurate models.  However, our study dataset is much smaller than the training data used to train modern state-of-the-art super-resolution models.  We have found that for our task and dataset SRCNN performs very well. In Table~\ref{tab:tab2}, we calculated the PSNR for the cross-validation image reconstruction task using the SRCNN network. Table~\ref{tab:model-comparison} shows the size and performance of five neural networks SRCNN, VGGExt, ResNetExt, SRDB and RCAN using the same dataset considered in Table~\ref{tab:tab2}.  The networks that performed best were SRCNN and SRDB.  We also tested how well these models generalizes on images of different crops.  Table~\ref{tab:out-of-domain-test}
 shows the results on images of wheat, corn, and miscanthus, and we see that the performance is quite good, so we expect that the spectral extension model can be applied to at least these other crops and to other locations in the northeast. SRCNN and SRDB are once again the better performing networks, but the other networks are not far behind.
 
RCAN is only matched by models that resort to generative adversarial network (GAN) architectures.  GAN models can be tricky to train and would not be possible to train for the modest dataset in our study. The VGG and ResNet architectures allow for deeper networks through the use of max-pooling and residual connections.  However, VGG and ResNet are primarily focused on image classification, and max-pooling can be harmful for super-resolution.  We constructed smaller versions of VGG and ResNet meant for our task with only one application of max-pooling that is paired with a conjugate convolution for upsampling later in the network.  The VGGExt and ResNetExt architectures can be seen in Figures~\ref{fig:VGGExt} and \ref{fig:ResNet}, respectively.  We also tried a smaller version of the Densenet architecture with no max-pooling layers that contain only the single dense block component shown in Figure~\ref{fig:SRDB}.  This super-resolution dense block (SRDB) network is similar in spirit to another state-of-the-art neural network, the residual dense network (RDN), which utilizes many dense blocks in a very deep architecture.  

\begin{table}[ht]
\caption{Comparison of SRCNN, VGGExt, ResNetExt, SRDB and RCAN on the study cross-validation data set.  The statistical significance of the numbers is approximately 0.2 DB, so the top two models (SRCNN and SRDB) should be considered comparable.}
\label{tab:model-comparison}.
\centering
\setlength\tabcolsep{2.0pt}
\begin{tabular}{lccccccc}
\toprule
      & Model: & SRCNN  & SRDB & VGGExt & ResNetExt & RCAN & RDN\\
      &  Size: & \textbf{0.52MB}  & 2.04MB & 3.31MB & 16.25MB & 15.23MB & 2.29MB\\
Field  & Date & \multicolumn{6}{c}{\texttt{PSNR}}\\
\cmidrule(lr){3-8}
\texttt{A}    & 3/20/19     & 35.69  & \textbf{35.70} & 34.87   & 33.74       & 32.43 & 32.86\\
              & 4/16/19     & 36.56  & \textbf{36.77} & 35.85   & 35.22       & 35.53 & 35.40\\
\texttt{B}    & 3/20/19     & 27.64  & \textbf{29.78} & 25.05   & 28.67       & 26.26 & 26.83\\
\texttt{I}    &             & \textbf{32.67}  & 32.29 & 30.71   & 32.01       & 30.49 & 29.95\\
\texttt{D}    &             & \textbf{32.61}  & 30.44 & 30.73   & 32.05       & 30.01 & 29.40\\
              & 4/16/19     & 29.08  & \textbf{30.53} & 29.90   & 29.51       & 29.22 & 28.33\\
\texttt{E}    & 3/20/19     & \textbf{32.33}  & 30.77 & 30.96   & 31.93       & 27.72 & 28.68\\
              & 4/16/19     & 32.82  & \textbf{34.71} & 31.87   & 31.50       & 33.84 & 33.29\\
\cmidrule(lr){3-8}
Average &         & 32.42  & \textbf{32.62} & 31.24   & 31.83       & 30.69 & 30.59\\
\bottomrule
\end{tabular}
\bigskip
\end{table}

\begin{table}[ht]
\caption{Comparison of the SRCNN, VGGExt, ResNetExt, SRDB and RCAN models trained on the Maryland cover crops dataset, but evaluated on a dataset with different crops (wheat, corn, miscanthus/switchgrass). This dataset contains 9 locations, but the images are larger and the total number of pixels is 10 times more than for the study.}\label{tab:out-of-domain-test}
\centering
\setlength\tabcolsep{2.0pt}
\begin{tabular}{lllcccccc}
\toprule
& & Model:  &SRCNN & SRDB & VGGExt & ResNetExt & RCAN & RDN \\
      &  & Size: &\textbf{0.52MB} & 2.04MB & 3.31MB & 16.25MB & 15.23MB & 2.29MB\\
Site  & Date & Crop & \multicolumn{6}{c}{\texttt{PSNR}}\\
\cmidrule(lr){4-9}
\texttt{W:A}      & 4/15/20 & wheat & \textbf{36.65} &  33.57 &  32.79 &  32.83 & 34.68 & 33.84\\
\texttt{W:B}       &         &       &  34.93 & \textbf{36.62}  &  31.18 &  32.16 & 35.15 & 35.62\\
\texttt{W:C}       &         &       &  \textbf{29.13} &  25.27 &  29.01 &  28.82 & 25.58 & 24.59\\
\texttt{W:D}       &         &       &  32.35 &  27.85 &  \textbf{34.29} &  31.90 & 29.04 & 26.85\\
\texttt{W:E}	 &         &	   &  \textbf{32.06} &  31.46 &  31.21 &  31.70 & 30.86 & 26.53\\
\texttt{W:F}    & 4/17/20 &	     &  29.68 &  29.47  &  30.97 &  29.55 & 27.67 & \textbf{31.52}\\
\texttt{C:A}  & 8/3/22  &  corn     &  28.81 &  \textbf{31.16} &  21.10 &  28.54 & 29.38 & 24.24\\
           & 7/29/20  &       &  25.90 &  26.79 &  29.53 &  25.79 & 25.51 & \textbf{31.10}\\
\texttt{C:B}   & 7/8/24  &   &  25.80 &  \textbf{30.05} &  21.39 &  29.31 & 29.98 & 29.92\\
M:A    & 6/8/20  & miscanthus &  30.33 &  31.62 &  \textbf{32.42} &  32.10 & 31.11 & 28.24\\ 
           & 9/23/20  &       &  \textbf{30.03} &  29.66 &  28.72 &  28.85 & 29.89 & 31.30\\
\cmidrule(lr){4-9}
Average    &         & wheat &  \textbf{32.47} &  30.71 &  31.57 &  31.16 & 30.50 & 29.79\\
           &         & corn  & 26.84  &  \textbf{29.33} &  24.00 &  27.88 & 28.29 & 28.56\\
           &         & miscanthus & 30.18  &  \textbf{30.64} &  30.57 &  30.47 & 30.50 & 29.67\\
           &         & all   & \textbf{30.52}  &  30.32 &  29.33 &  30.14 & 30.03 & 29.43\\
\bottomrule
\end{tabular}
\bigskip
\end{table}

\subsubsection{Spectral extension on cloudy days}
Also important when considering using the spectral extension model for real applications is whether cloud-free satellite imagery is even available.  In Table~\ref{tab:srcnn-rgb} we compare an SRCNN model trained using 11 input channels (8 Sentinel-2 bands and 3 UAS RGB bands) versus an SRCNN model trained only on UAS RGB data.  Unsurprisingly, the model that used the additional Sentinel-2 bands performed better by about 1.6 dB.  However, for two images, the RGB based model actually performed slightly better.  This is a testament to the importance of the RGB data.  We also had access to UAS data from 11/20/2019 where there was no temporally adjacent cloud-free Sentinel-2 imagery.  Before 11/20 there were no cloud-free Sentinel-2 images after 11/2 and after 11/20 there were no cloud-free Sentinel-2 images before 12/12.  This is therefore an opportunity to test the RGB only model.  Table~\ref{tab:srcnn-fall} shows the performance of the model with UAS RGB input only compared to the model that uses Sentinel-2 data from 11/2 and 12/12, respectively.  In this case the RGB only model outperforms the competition, but even with such mismatched Sentinel-2 data the competing models are not far behind.



\begin{table}[ht]
\caption{A comparison of spectral extension SRCNN models utilizing UAS RGB input only versus also taking advantage of Sentinel-2 data.  A model relying solely on UAS RGB input is particularly useful in situations where no cloud-free Sentinel-2 data is available. } \label{tab:cloudy}
\begin{subtable}[t]{0.45\textwidth}
\subcaption{Comparison when recent cloud-free Sentinel-2 data is available.}\label{tab:srcnn-rgb}
\centering
\begin{tabular}{lccc}
\toprule
      & Target Sentinel-2 & Yes & No  \\
Field  & Date & \multicolumn{2}{c}{\texttt{PSNR}}\\
\cmidrule(lr){3-4}
\texttt{A}    & 3/20/19     & 35.69  & 32.47\\
              & 4/16/19     & 36.56  & 30.89\\
\texttt{B}    & 3/20/19     & 27.64  & 27.06\\
\texttt{I}    &             & 32.67  & 29.96\\
\texttt{D}    &             & 32.61  & 31.95\\
              & 4/16/19     & 29.08  & 28.78\\
\texttt{E}    & 3/20/19     & 32.33  & \textbf{32.59}\\
              & 4/16/19     & 32.82  & \textbf{33.08}\\
\cmidrule(lr){3-4}
Average &         & \textbf{32.42}  & \textbf{30.85}\\
\bottomrule
\end{tabular}
\end{subtable}\hfill
\begin{subtable}[t]{0.5\textwidth}
\subcaption{A situation where no recent Sentinel-2 data was available.  For the UAS flights on 11/20/19 and 11/21/19 the closest available cloud-free Sentinel-2 images were on 11/2 and 12/12. }\label{tab:srcnn-fall}
\centering
\begin{tabular}{lcccc}
\toprule
      & Target Sentinel-2 &  None & 11/2/19 & 12/12/19 \\
Field  & Date & \multicolumn{3}{c}{\texttt{PSNR}}\\
\cmidrule(lr){3-5}
\texttt{B}    & 11/20/19     & 27.98 & \textbf{30.67} & 23.62 \\
\texttt{H}    &     & 28.34 & \textbf{29.75} & 25.46 \\
\texttt{E}    & 11/21/19   & \textbf{29.39} & 23.78 & 26.56 \\
\cmidrule(lr){3-5}
Average &         & \textbf{28.57} & 28.07 & 25.21 \\
\bottomrule
\end{tabular}
\end{subtable}
\end{table}

\subsection{Experimental results of super-resolution fusion approaches for precision management}

Table \ref{tab:tab3} shows the results of the cover crop biomass yield and quality estimation of our RF models on the reconstructed high-resolution MS images. These experiments aimed to reproduce the improvements for the biomass and N estimates that we observed in Section~\ref{sec:spectral_performance}. The RF model \modelref{spectral:SRCNN}, which used reconstructed MS bands (from the Spectral-SRCNN model) as predictors, performed substantially better than the RF models (\modelref{spectral:S2}, \modelref{spectral:S210bands}) that used predictors from the Sentinel-2 benchmark data (ie, $\approx18\%$ reduction in RMSE for biomass; $\approx31\%$ reduction in RMSE for N). 

In terms of improvement over flying a high spatial resolution RGB UAS (\modelref{hyp:S2:RGB} in Table \ref{tab:org_models} and \modelref{spectral:S2:RGB}), our approach (\modelref{spectral:SRCNN}) performed slightly better for biomass yield estimation and substantially improved performance ($\approx31\%$ reduction in the MSE) for N estimation, suggesting that both spatial resolution (i.e., how much plant is there) and spectral range matter for biomass prediction, and the spectral range matters more to N prediction. In other words, both leaf area and plant chemistry matter for biomass prediction, whereas plant chemistry is more important for N content prediction. 

\begin{table}[h!]
\caption{Impacts of reconstructed spatial and spectral resolution and spectral range on RF model performance for biomass yield and N content estimations.
Note that the test data differs as follows: Spectral: sites A, D, E in April and March,  sites B and I in  March; Spatial: sites A, E in March and April, Temporal: sites A, D, E in December only.  We abbreviated ground-truth as GT.}
\centering
\footnotesize
\setlength{\tabcolsep}{4pt} 
\begin{tabular}{lllcccc}
\toprule
Extension & Tag & Data Source (ground-truth=GT)& \multicolumn{2}{c}{Biomass} & \multicolumn{2}{c}{Nitrogen}\\
Scenario &  &  & $R^2$      & RMSE (Mg/ha)  & $R^2$      & RMSE (kg/ha)   \\
\cmidrule(lr){4-5}\cmidrule(lr){6-7}
Spectral & \modellabel{spectral:S2} & Sentinel-2 8-band (10m)                          
         & 68.9    & 0.87 & 55.9 & 25.1 \\
         & \modellabel{spectral:S210bands} & Sentinel 10-band (with SWIRs) (10m)             
         & 65.9    & 0.91 & 53.4 & 25.7 \\         &\modellabel{spectral:S2:RGB} 
         & UAS RGB (0.125m) & 80.7    & 0.72        & 59.9     & 25.1\\
         & \modellabel{spectral:UAS8} & \textit{UAS 8-band (0.125m)}                       
         & \textit{84.5}    & \textit{0.67} & \textit{82.0} & \textit{17.7} \\
         & \modellabel{spectral:SRCNN} & \textbf{Spectral-SRCNN (8-fold) 8-band (0.125m)}           
         & \textbf{81.9}    & \textbf{0.71} & \textbf{83.7} & \textbf{17.2} \\

\cmidrule(lr){2-3}
Spatial  & \modellabel{Spatial:S2} 
         & Sentinel-2 8-band (1m) & 81.8    & 0.78 & 82.1 & 20.5 \\
         & \modellabel{Spatial:UASRGB} 
         & UAS RGB (0.125m) & 88.5    & 0.65 & 87.7 & 17.8 \\         & \modellabel{Spatial:UAS8} 
         & \textit{UAS 8-band (0.125m)} & \textit{93.2}    & \textit{0.49} & \textit{93.6} & \textit{12.6} \\
         & \modellabel{Spatial:SRCNN:fromhyp} 
         & Spatial-SRCNN (a) 8-band (1m), GT: UAS 8-band            
         & 88.1    & 0.66 & 86.8 & 18.3 \\
         & \modellabel{Spatial:SRCNN:fromspectral} 
         & \textbf{Spatial-SRCNN (b) 8-band (1m) GT: Spec-SRCNN}           
         & \textbf{89.8}    & \textbf{0.6} & \textbf{88.7} & \textbf{16.7} \\

\cmidrule(lr){2-3}
Temporal & \modellabel{temporal:S2} 
         & Sentinel-2 8-band (1m)                         
         & 38.7    & 7.05 & 49.3 & 177.1 \\
         & \modellabel{temporal:STSRCNN}
         & \textbf{ST-SRCNN 8-band (1m)}                          
         & \textbf{58.3}    & \textbf{6.84} & \textbf{61.5} & \textbf{168.6} \\

\bottomrule
\end{tabular}
\label{tab:tab3}
\end{table}

When the farmer needs to generate datasets on a portion of their farm without flying, we resort to the Spatial-SRCNN model. However, the Spatial-SRCNN does not generalize as well as the Spectral-SRCNN, so the farmer needs a specialized Spatial-SRCNN unless he grows the same crop in the same geographic area (in which case our model can be used). For this, there are two scenarios: the farmer obtains ground-truth data using hyperspectral UAS data or use the Spectral-SRCNN to predict the ground-truth data. The last scenario is possible, since Spectral-SRCNN generalizes better than Spatial-SRCNN. Both RF models \modelref{Spatial:SRCNN:fromhyp} and \modelref{Spatial:SRCNN:fromspectral} with reconstructed MS predictors from the corresponding spatial extension SRCNNs (a \& b) performed better than the Sentinel-2 benchmark (\modelref{Spatial:S2}). \modelref{Spatial:S2} shows RF model results over the non-flown areas (i.e., where UAS data is not available) using the resampled Sentinel-2 8-band data (1 m) as predictors. The difference between Spatial-SRCNN (a) and Spatial-SRCNN (b) lies in their use of ground-truth data in the model training; the former used the UAS 8-band data as ground-truth, whereas the latter used our Spectral-SRCNN inferred high-resolution data as ground-truth. Thus, Spatial-SRCNN (b) allows the farmer to train a specialized spatial extension model when only an RGB camera is available that can be used to generate the high-resolution ground-truth data required to train a Spatial-SRCNN model.    

Spatial-SRCNN (a) improved over \modelref{Spatial:S2} by $\approx15\%$ reduction in MSE for biomass and $\approx11\%$ reduction in MSE for N, while Spatial-SRCNN (b) improved over \modelref{Spatial:S2} by $\approx23\%$ reduction in MSE for biomass and $\approx19\%$ reduction in MSE for N. This means that in terms of estimating biomass and N, substantial improvements can be made if the UAS 8-band ground-truth is available to train a Spatial-SRCNN model. However, if only the RGB camera data are available for training a Spatial-SRCNN (b), the performance is still better than \modelref{Spatial:S2}  and \modelref{Spatial:SRCNN:fromhyp} . Moreover, \modelref{Spatial:UASRGB} results suggest that by using this Spatial-SRCNN (b) approach the biomass and N predictions are better than predicting from actual UAS RGB data. Thus, the farmer can stop flying UAS once a specialized spatial extension model has been trained from the targeted UAS RGB data.

The RF model \modelref{temporal:STSRCNN} with reconstructed MS predictors from the ST-SRCNN model performed better than the Sentinel-2 benchmark (16), highlighting its promise in scenarios when the farmer needs to generate data over their fields from a different temporal period. \modelref{temporal:STSRCNN} improved over \modelref{temporal:S2} by reducing $\approx3\%$ MSE for biomass and $\approx4.8\%$ for N estimation. Note that this experimental set-up applies to the conditions when UAS data is available, for example, for a farm over the past spring season but not available for the current or next one.

\section{Discussion}
\label{sec:discussion}
Our study investigated how UAS images (RGB, multispectral, or hyperspectral) compare with Sentinel-2 data to monitor cover crop health, specifically in terms of biomass yield and N content at the farm scale. Building on these insights, we developed a novel approach to sharpen Sentinel-2 imagery and spectrally extend UAS RGB data. We then examined how these reconstructed RS datasets can improve cover crop modeling across spectral, spatial, and temporal domains. Specifically, we explored how the proposed methods generalize across time periods and agricultural fields, particularly under constraints such as limited UAS access or persistent cloud cover. Our findings indicate that effective monitoring of cover crop health depends not only on spatial resolution but also on the spectral range and resolution of multispectral RS imagery. While high-resolution UAS imagery with VNIR spectral coverage consistently outperformed Sentinel-2, the high cost of UAS deployment remains a limiting factor in precision farming. To address this, we demonstrated that UAS-guided super-resolution approaches can yield significant improvements in predictive modeling across spectral, spatial, and temporal dimensions, offering a cost-effective strategy for advancing precision agriculture. 

Our findings are consistent with previous studies that reported a positive association between spatial resolution and biomass yield estimation accuracy \cite{assmann2020drone, salehin2025cover}. Similarly, our results on cover crop nitrogen estimation accuracy align with those that used narrow MSI spectral bands (15 nm bandwidth) centered at 705 nm and 740 nm, such as red-edge bands, to estimate crop and grass chlorophyll and N content \cite{clevers2013remote}, N uptake in maize crop \cite{sharifi2020using}, and canopy N content in rice cropping system \cite{rossi2023sentinel}. These confirm that an extended and higher spatial resolution spectral range beyond RGB range is important for accurate nitrogen estimation. These findings support the earlier premise that a farmer with a hyperspectral sensor covering the 400-2,500 nm range would achieve the optimal model performance. Our analysis with hyperspectral data further suggests that a spectral resolution of about 100 nm is sufficient for accurate estimation of both biomass and N, which is comparable to the spectral distance between different Sentinel-2 bands. However, collecting high-resolution imagery across large areas using UAS platforms remains impractical due to operational and cost constraints. Therefore, as we have shown via super-resolution fusion strategies, the synergy of UAS and Sentinel-2 imagery at spatial, temporal, and spectral dimensions offers an alternative approach to achieve improved model performance for precision agriculture. In particular, spectral extension allows farmers to use low-cost UAS platforms equipped with RGB cameras while still leveraging the broader spectral range of Sentinel-2 or other satellites without requiring expensive multi- or hyperspectral sensors. Unlike spatial and temporal extension models, the spectral extension model requires both UAS and satellite imagery but retains the high spatial resolution of the original UAS data. As a result, the method produces sharpened and spectrally enriched Sentinel-2 MS bands, enabling more detailed field-level analysis management practices.

\begin{figure}[h!]
  \centering
  \includegraphics[width=0.65\textwidth]{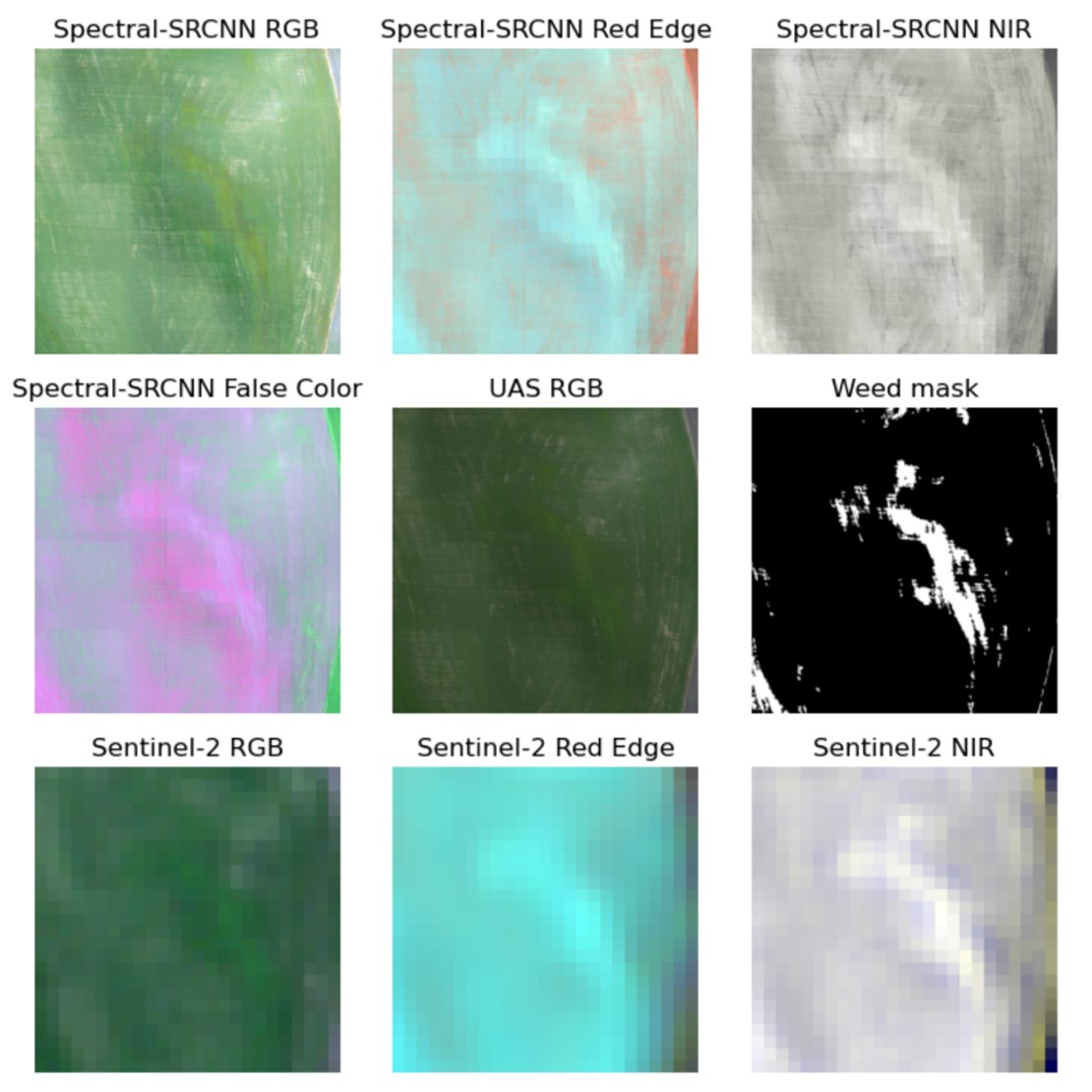}
  \caption{The figure shows a UAS image from a winter wheat field in Washington state in the center surrounded by Spectral-SRCNN, an annotated weed mask and Sentinel-2 images from the same location. The false color composite uses Narrow band NIR (NNIR), Green and VRE 2. These bands were chosen to enable the annotated streak of Italian rye grass to pop out in the image.  The streak appears in bright pink/magenta in the false color image and less contrastive as a lighter green in the UAS image.}
  \label{fig:fig9}
\end{figure}

Another application we demonstrated in this study is the identification of weed infestations. When only RGB imagery is available, the weed patch --- slightly brighter green hue --- can be difficult to distinguish (Figure~\ref{fig:fig9}). However, the inclusion of sharpened VRE and NIR bands makes the weed patch clearly visible, without the need for a costly multi- or hyperspectral sensor. This application is analogous to the well-known use of UV light to reveal otherwise invisible contamination, highlighting the potential of spectral extension to uncover features not detectable with conventional RGB imagery alone \cite{li2021identification}.  

As with spectral extension, targeted UAS imagery can also be utilized for spatial and temporal extension models. The spatial extension model allows the farmer to fly a small and representative subset of their fields, and this high-resolution imagery is leveraged to provide information to the SRCNN model to interpret the Sentinel-2 imagery. For temporal extension, UAS images collected at different time points help improve the alignment and interpretation of Sentinel-2 data as crops progress through various phenological stages. Although weekly UAS flights may be beneficial to the model, this frequency may not be necessary depending on the type of precision crop management decisions that are considered. For N management, for example, decisions are made within a defined narrow range of crop development \cite{zhang2015managing}. Although the UAS imagery collection can be crop- and context-specific (e.g., responsive phenological stages, stress detection, managing variable-rate nitrogen application, etc.) data collected in one season can be leveraged in subsequent years, thus requiring fewer images over time. 

This strategy of using targeted UAS imagery across space, time and spectrum contrasts with other studies that have fused UAS and satellite imagery with the same spatial and temporal extent \cite{brook2020smart}. Full-coverage fusion can be effective but is impractical from a farmer’s perspective due to operational and cost constraints, as discussed earlier. Contrary to using image fusion at a common spatial extent, we have shown that targeted UAS imagery fused with satellite imagery using SRCNN can achieve many of the same objectives at a sub-meter spatial resolution, which may be consistent with many requirements for precision crop management, such as scouting for problems, monitoring to prevent yield losses, and planning crop management operations \cite{hunt2018good}. When only Sentinel-2 data are available, our models produce images at a resolution of 1 m. In overlapping areas where both UAS and Sentinel-2 data are available, we achieve sub-meter resolution (0.125 m) with no apparent loss in fidelity. In this fusion context, our Spectral-SRCNN model is novel in that it generates high-resolution Sentinel-2 data using hyperspectral targets with minimal loss of fidelity, which laid the foundation for the strong performance observed in the context of cover cropping practices.

The superior performance of the spectral extension model generated reconstructed high-resolution imagery in cover crop application is not unexpected, given that the model had access to both high-resolution RGB input and wide-range spectral input from Sentinel-2. The spectral bands for Sentinel-2 are designed for agricultural and environmental monitoring, and the spectral extension model achieves performance that closely approaches that of models based on hyperspectral UAS data. The spectral extension model also generalizes well to other regions and crop species, and although there is some loss in performance, the generated images can still serve as ground-truth data for a farmer that needs to train a spatial extension model specific to a region or crop. This scenario deserves consideration because high-quality hyperspectral sensors can cost as much as \$175,000 \cite{olson2021review}, and the spatial extension model exhibits reduced generalization compared to the spectral extension model. If there is only one target application, it may make sense to build a task-specific neural network. However, the benefit of harmonizing hyperspectral images with a satellite sensor is that we can take advantage of the models available for that satellite, such as vegetation indices and random forest models. In the case of Sentinel-2 this immediately gives us accessibility to a plethora of models.

Building on these proposed and validated multi-dimensional super-resolution capabilities, an AI-driven alert system could be developed to provide early warnings to farmers of potential anomalies developing in their fields due to emerging weed infestations, insect outbreaks, or disease symptoms. Although satellite imagery alone may lack the much needed spatial resolution and UAS may be limited in spatial and spectral coverage, their combination offers both improved resolution and sensitivity to early and more accurate detection of these field-level anomalies. A passive alert system could provide sufficient early warning itself or prompt targeted in-field scouting for further investigation. 

\subsection{Trade-off between model accuracy and cost to the farmer}
\begin{figure}[ht]
  \centering
  \includegraphics[width=0.8\textwidth]{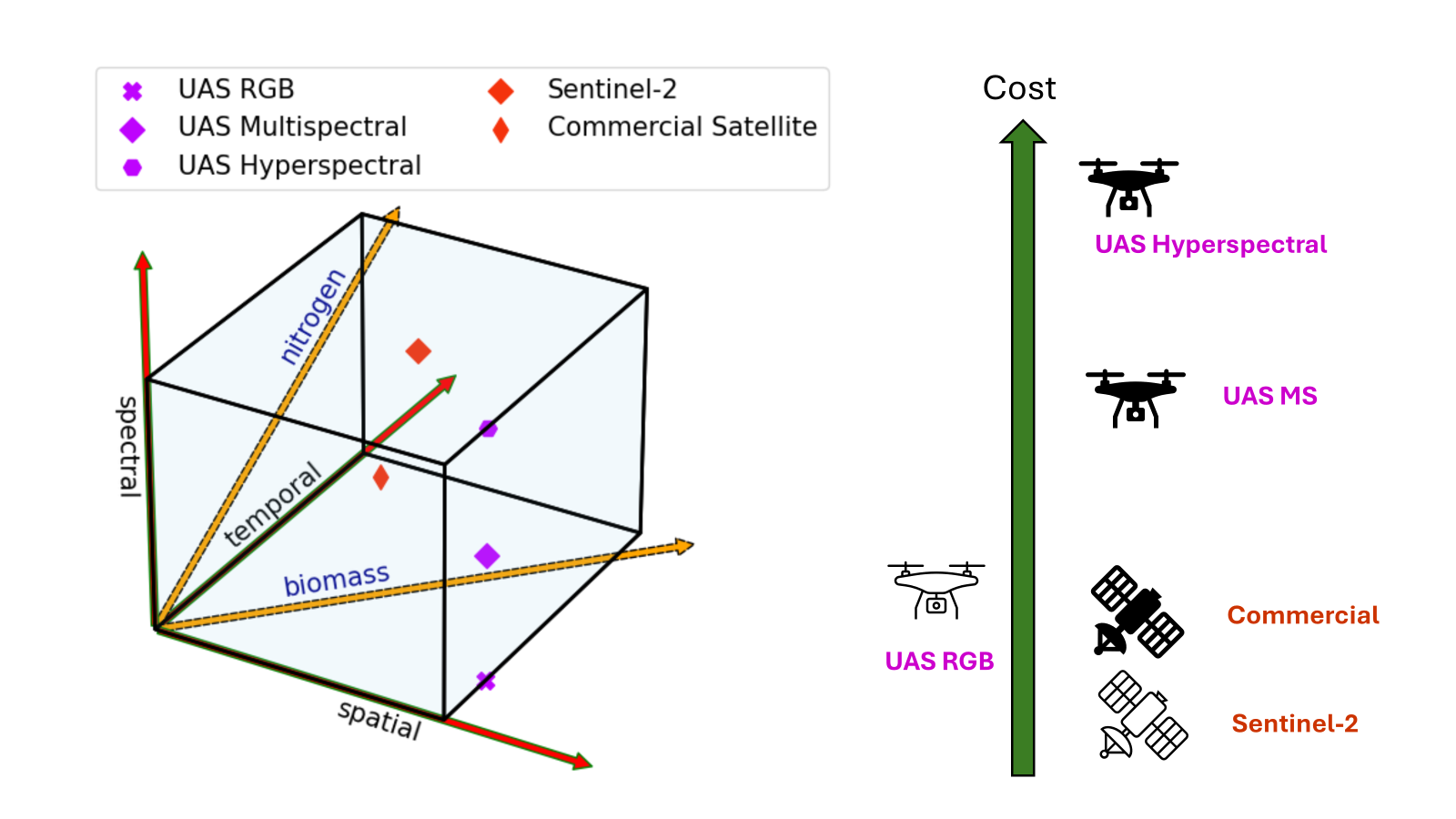}
  \caption{Effectiveness of different sensors for super-resolution across spectral, spatial, and temporal domains in estimating biomass yield and N content and their cost-effectiveness.}
  \label{fig:fig10}
\end{figure}

\textbf{Cost Considerations:} From a farmer’s perspective there is a trade-off between application accuracy and obtaining UAS imagery. Regulations for flying a UAS vary greatly by country \cite{UAS_Regulations_FAA}. However, any commercial operator in the US requires authorization in a Part 107 Certificate. This knowledge gap leads fewer farmers to operate their own UAS and instead resort to obtaining it through a service. The UAS pilot rates for agricultural services were reported to be on average \$162/hour according to \cite{simula2021establishing}. On top of the pilot rates there is the additional cost of managing the imagery and these expenses are proportional to the number of flights. Moreover, \cite{olson2021review} gives the price of an MS camera as much as \$10 K and \$175 K for a hyperspectral camera, thus making it significantly more expensive to obtain spectral information beyond the RGB range. Figure~\ref{fig:fig10} highlights cost-effectiveness scenarios for our applied use case of estimating biomass yield and N content from different UAS and satellite datasets. 

\begin{figure}[h!]
  \centering
  \includegraphics[width=0.65\textwidth]{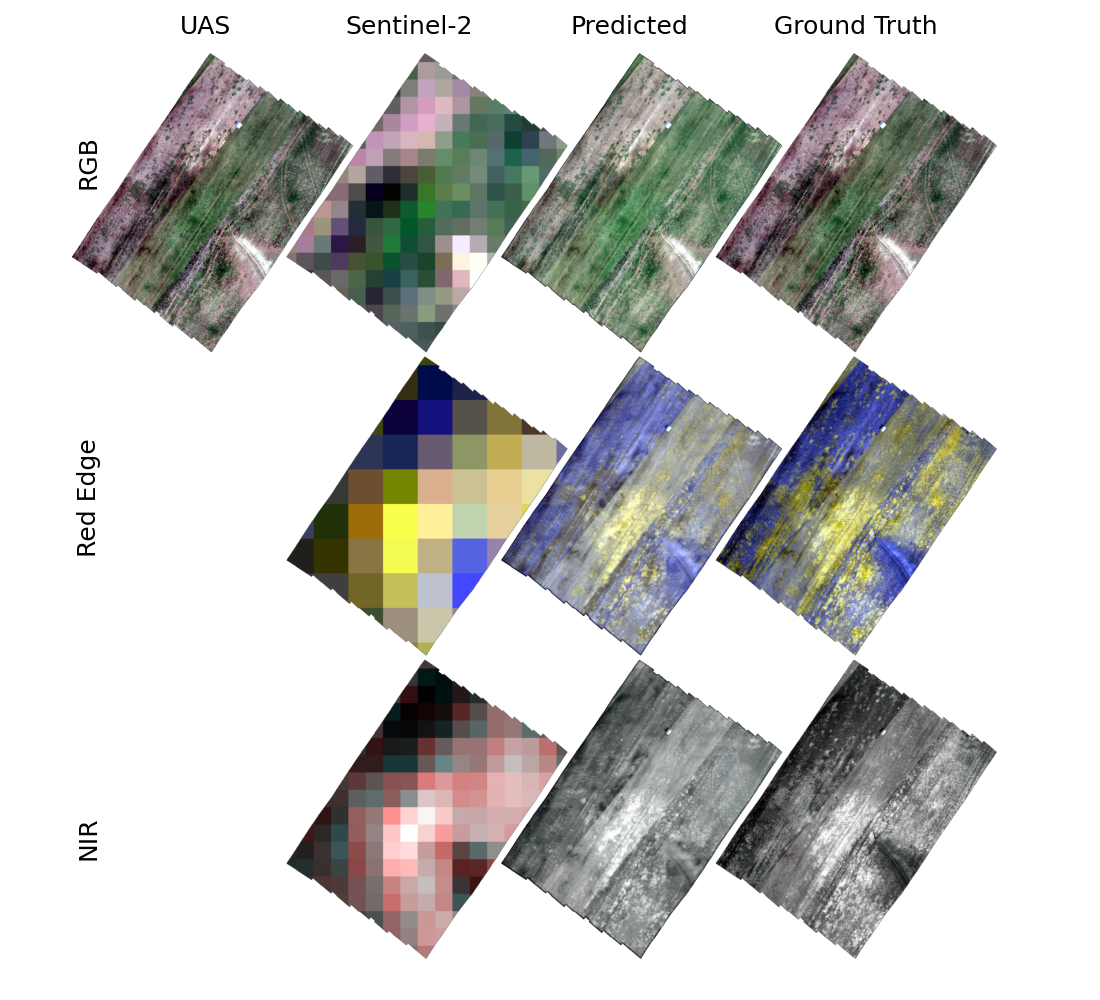}
  \caption{The spectral extension model compared to the ground-truth generated by the hyperspectral camera for site I.  The first two columns represents the input to the model and the third column is the output from model \modelref{spectral:SRCNN}.}
  \label{fig:spectral_extension_visualized}
\end{figure}

\begin{figure}[ht]
  \centering
  \includegraphics[width=0.65\textwidth]{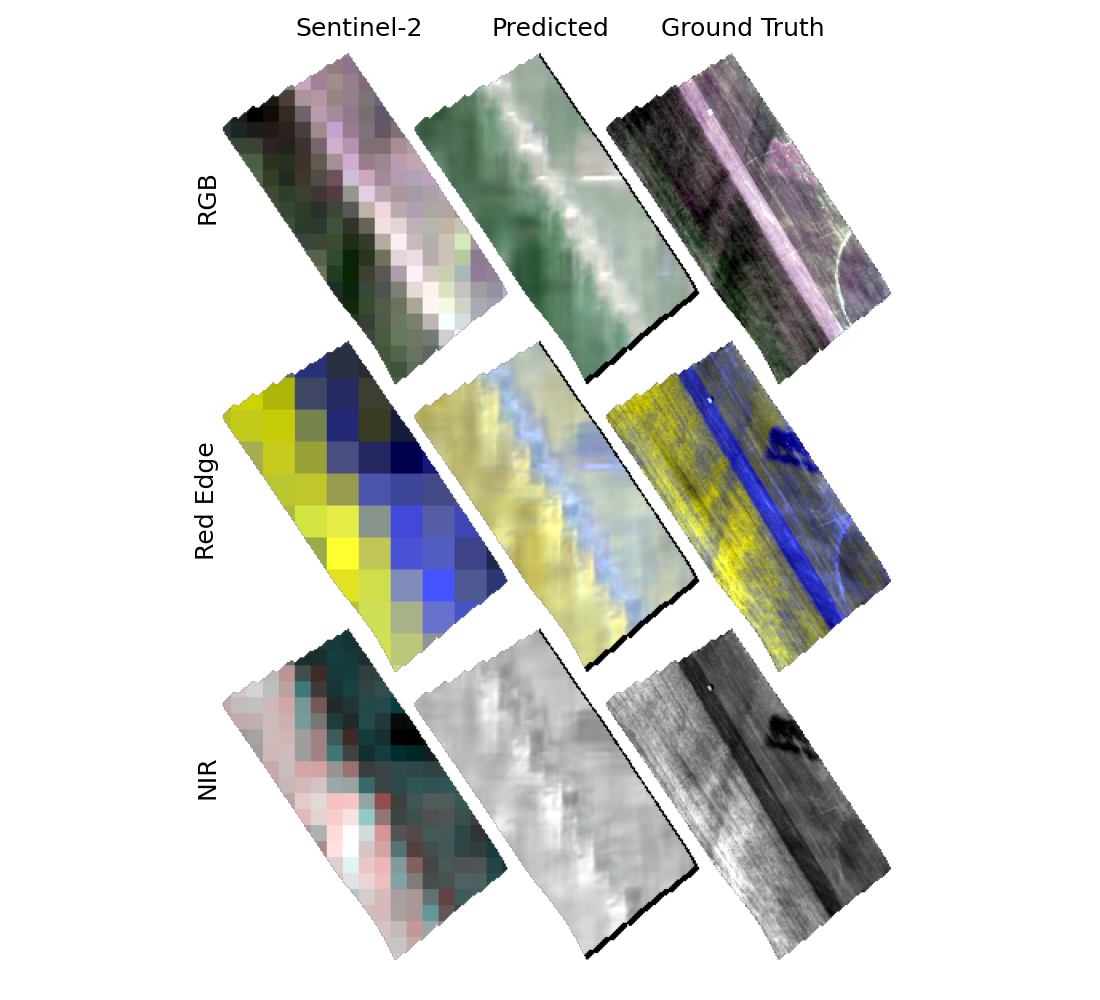}
  \caption{The spatial extension model \modelref{Spatial:SRCNN:fromspectral} compared to the ground-truth generated by the hyperspectral camera for site E.  The first column is the Sentinel-2 input to the model, while the second column is the super-resolved prediction and the third column is the desired ground-truth.}
  \label{fig:spatial_extension_visualized}
\end{figure}

\textbf{Accuracy Considerations:} We have already seen in Table~\ref{tab:org_models} that different applications have different requirements in terms of spatial and temporal resolution and spectral range and resolution. For example, the N content prediction improves with greater spectral range and resolution in contrast to the biomass yield prediction, and therefore a higher quality UAS camera may be beneficial (Figure~\ref{fig:fig10}). In the case where the spectral bands of the Sentinel-2 satellite are sufficient, we have shown that an RGB camera together with a cloud-free Sentinel-2 image can suffice. Very little degradation is observed compared to the same sensor observations obtained using a UAS, as can be seen in Table~\ref{tab:tab2} and Figure~\ref{fig:spectral_extension_visualized}. We have little evidence to assess how well the super-resolution methods will generalize outside of the location and cover crop setting. However, as seen in Figure~\ref{fig:fig9} and Table~\ref{tab:out-of-domain-test} there is some evidence that the spectral extension model will generalize, while the lower performance of the temporal and spatial extension models is an indication that these will have to be re-trained or trained on a substantially larger training dataset. As we see in Figure~\ref{fig:spatial_extension_visualized} the quality of the spatial extension model is not at the level of the spectral extension model.  This is understandable as there is no longer any high-resolution side information given to the model and as a result the spatial extension model generalizes less well and may have to be retrained for different crops or locations. The question from the farmers viewpoint would then be whether retraining these models requires an actual hyperspectral camera or whether the spectral extension model could suffice to generate synthetic ground-truth data.  

\textbf{Trade-off:} We have shown that Sentinel-2 imagery could be improved using targeted UAS imagery. However, we do not know if this improvement will always be considered cost-effective or how much imagery is required to achieve performance gains, which are considered cost-effective. At one end, the farmer could choose not to fly any of the fields, depending only on Sentinel-2 imagery if performance gains are not sufficient. However, for the spectral extension model, we showed that they could use the trained model with little loss in model accuracy without the need for any UAS flights. We do not know how robust or how sensitive this model is to changes in geography and crop type. 

\section{Limitations}
\label{sec:limitations}

Although our analysis has shown that UAS and satellite imagery play a complementary role that can positively impact predictive model performance in precision management practices, there are some limitations that could be improved through further research. While the use of targeted UAS imagery with higher spatial and spectral resolution and range improves the Sentinel-2 model performance for cover crop estimations, it is yet unknown what the optimal extent of imagery coverage or temporal resolution is in the trade-off between performance gains and acquisition costs. We have shown that spatial extension with limited hyperspectral imagery is possible for cover crop management. It remains uncertain how far geographically this modeling approach will extend or whether geographical changes will affect the spectral, spatial, and temporal models differently. It may be required to develop one model for each available satellite platform. The lower the resolution of the satellite platform, the more data is needed. Since hyperspectral data collection is expensive this limits which satellite can be considered. Our current imagery comes from multiple counties across Maryland and Pennsylvania, but the models may generalize further geographically. However, our imagery has limited temporal coverage, which probably reduced model performance for the temporal extension. Covering multiple years and covering all stages of crop development may improve the temporal extension model.

Future investigations could include efforts to improve model performance using larger datasets. How much would broader temporal and spatial coverage improve the model's ability to generalize? In addition, different combinations of spectral bands and sensors should be explored. Further improvement of model performance could be possible by using other state-of-the-art super-resolution models such as SR3 (Super-Resolution via Repeated Refinement) \cite{saharia2021sr3}. It should be noted that there are some regulations that preclude the use of generative models such as SR3 and possibly even GAN-based models for use in government run programs.  Finally, development of an alert detection system that incorporates pre-trained fusion models and cover crop estimations, among other relevant applications, could benefit farmers with cost-effective and targeted management practices. 

\section{Conclusions}
\label{sec:conclusions}

This study demonstrated a scalable, end-to-end super-resolution system that integrates satellite and UAS imagery to support cost-effective precision farming practices. Although illustrated through a case study on winter cover crops in the Chesapeake Bay region, the methodology is broadly applicable to other crops, regions, and precision agriculture scenarios. By leveraging super-resolution CNN, we showed that spatial, spectral, and temporal limitations of each platform can be mitigated through targeted fusion. For example, the reconstructed sub-meter resolution Sentinel-2 images improved the estimation of biomass and N by up to 18\% and 31\%, respectively, compared to the 10 m resolution benchmark. This approach offers a practical solution for precision agricultural management that balances resolution, scalability, and affordability, without relying on expensive hyperspectral sensors. 

We have shown how targeted UAS data (both multispectral and hyperspectral), strategically collected at select locations and times, can be effectively integrated with satellite images to enhance spatial, spectral and temporal resolution while reducing operational costs. By flying a subset of fields with low-cost RGB cameras, farmers can extend the spectral range of these imagery to include critical VRE and NIR bands, and potentially SWIR bands if high-resolution ground-truth data are available, without needing any expensive sensors. Similarly, the spatial and temporal coverage of these enhanced images can be improved by leveraging the wide availability of satellite imagery. While extending the spatial coverage we have shown that the proposed spatial extension SRCNN model yields better biomass and N predictions than those from actual UAS RGB data. Thus, the farmer can stop flying UAS once a specialized spatial extension model has been trained from targeted UAS RGB data. Moreover, a SRCNN model that relies solely on the UAS RGB input can be useful in situations where there are no cloud-free Sentinel-2 data available. These flexible fusion strategies across space, time, and spectrum offer a scalable solution for precision crop management across varying crop types, field sizes, and growing conditions. Future research is needed to further assess the robustness of this super-resolution AI system across space and time. An early warning system can be developed that enables farmers to proactively and affordably address various anomalies in their agricultural fields, including emerging weed infestations, insect outbreaks, and disease symptoms.

\section*{Acknowledgments}
Mention of trade names or commercial products in this publication is solely for the purpose of providing specific information and does not imply recommendation or endorsement by the U.S. Department of Agriculture. USDA is an equal opportunity provider and employer. This research was a contribution from the Long-Term Agroecosystem Research (LTAR) network. LTAR is supported by the USDA. Not subject to copyright in the USA. Contribution of United States Department of Agriculture, Agricultural Research Service (USDA-ARS).

In addition to the support from USDA, the authors would also like to thank ESRI, Andrew Nelson (Nelson Wheat), Ranveer Chandra (Microsoft), Anirudh Badam (formerly Microsoft), and Roberto Estevão (Microsoft) for providing resources, data, and useful feedback.  Special thanks to Roberto Estev\~{a}o for adapting an early version of the spectral extension model for inclusion in the open source repository FarmVibes\footnote{\href{https://github.com/microsoft/farmvibes-ai/blob/main/notebooks/spectral_extension/spectral_extension.ipynb}{https://github.com/microsoft/farmvibes-ai/blob/main/notebooks/spectral\_extension/spectral\_extension.ipynb}}.

\bibliographystyle{apalike}  
\bibliography{references}  

\newpage
\begin{appendices}

\section{Appendix}

\renewcommand{\thefigure}{A.\arabic{figure}}  
\setcounter{figure}{0}  


\begin{figure}[ht]
  \centering
  \includegraphics[align=c, width=1.0\textwidth]{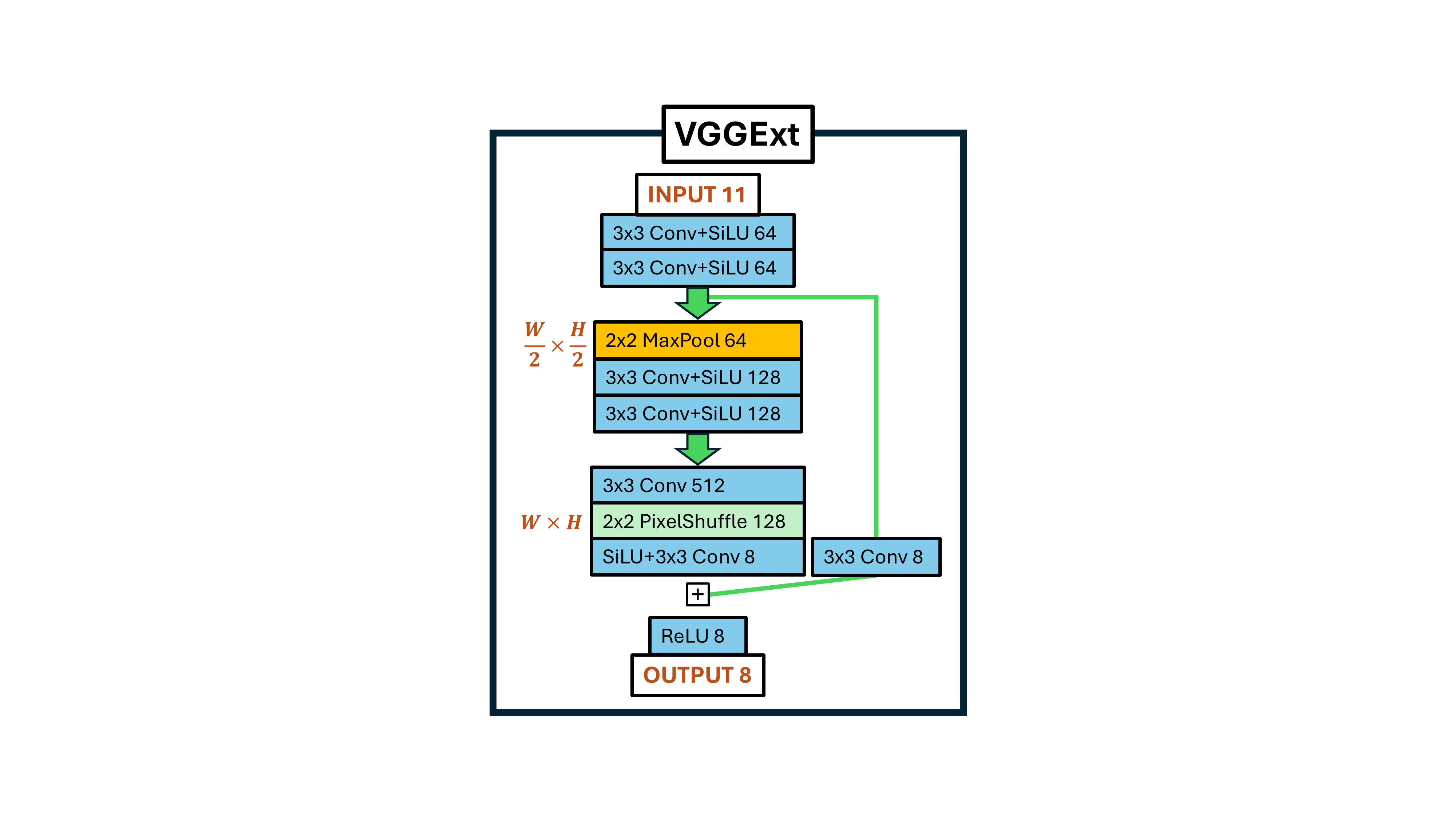}
  \caption{The VGGExt network architecture. We show the number of channels on the right and when needed the image size on the left.}
  \label{fig:VGGExt}
\end{figure}

\begin{figure}[ht]
  \centering
  \includegraphics[width=1.0\textwidth]{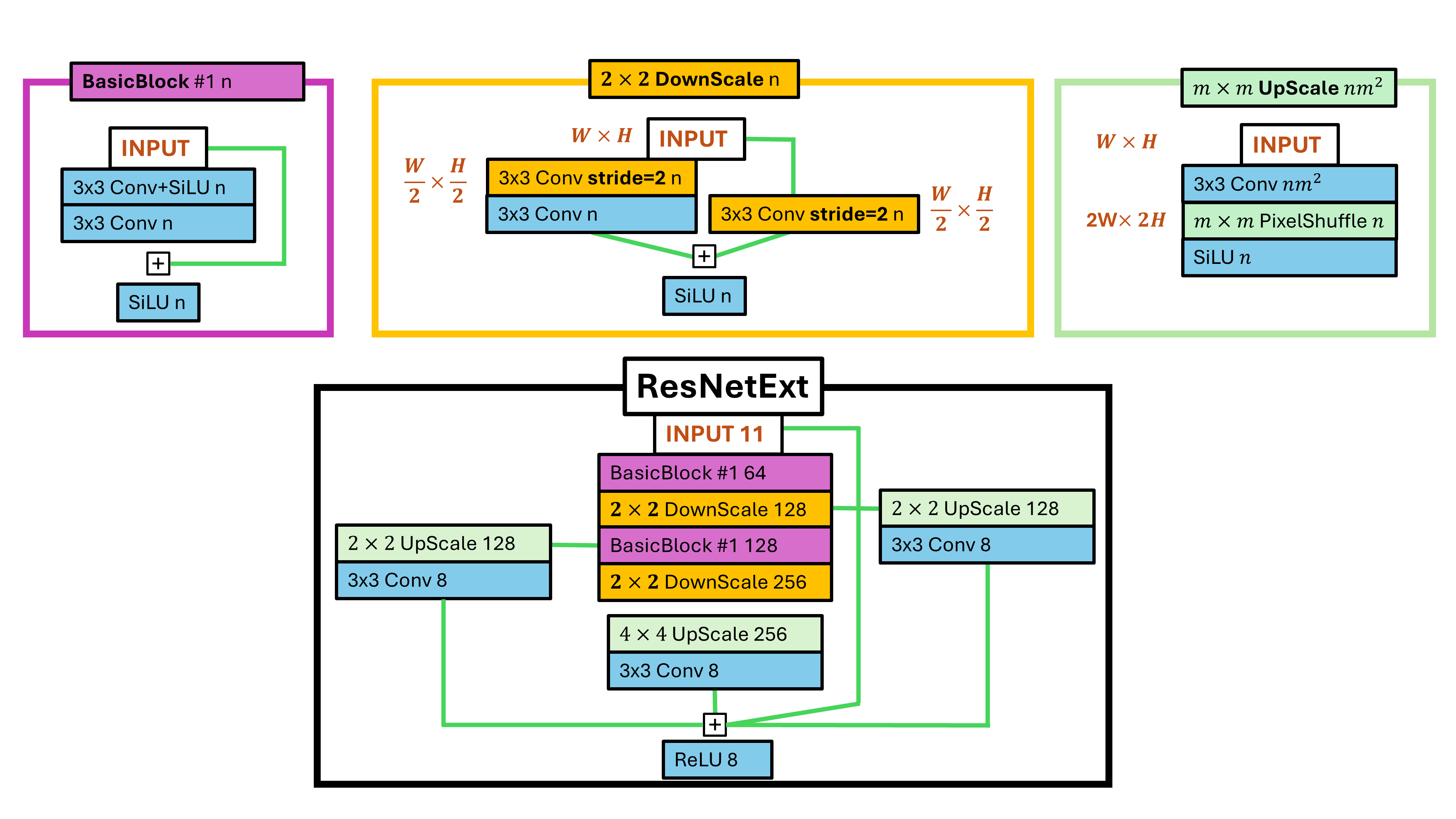}
  \caption{The ResNetExt network architecture.}
  \label{fig:ResNet}
\end{figure}

\begin{figure}[ht]
  \centering
  \includegraphics[width=1.0\textwidth]{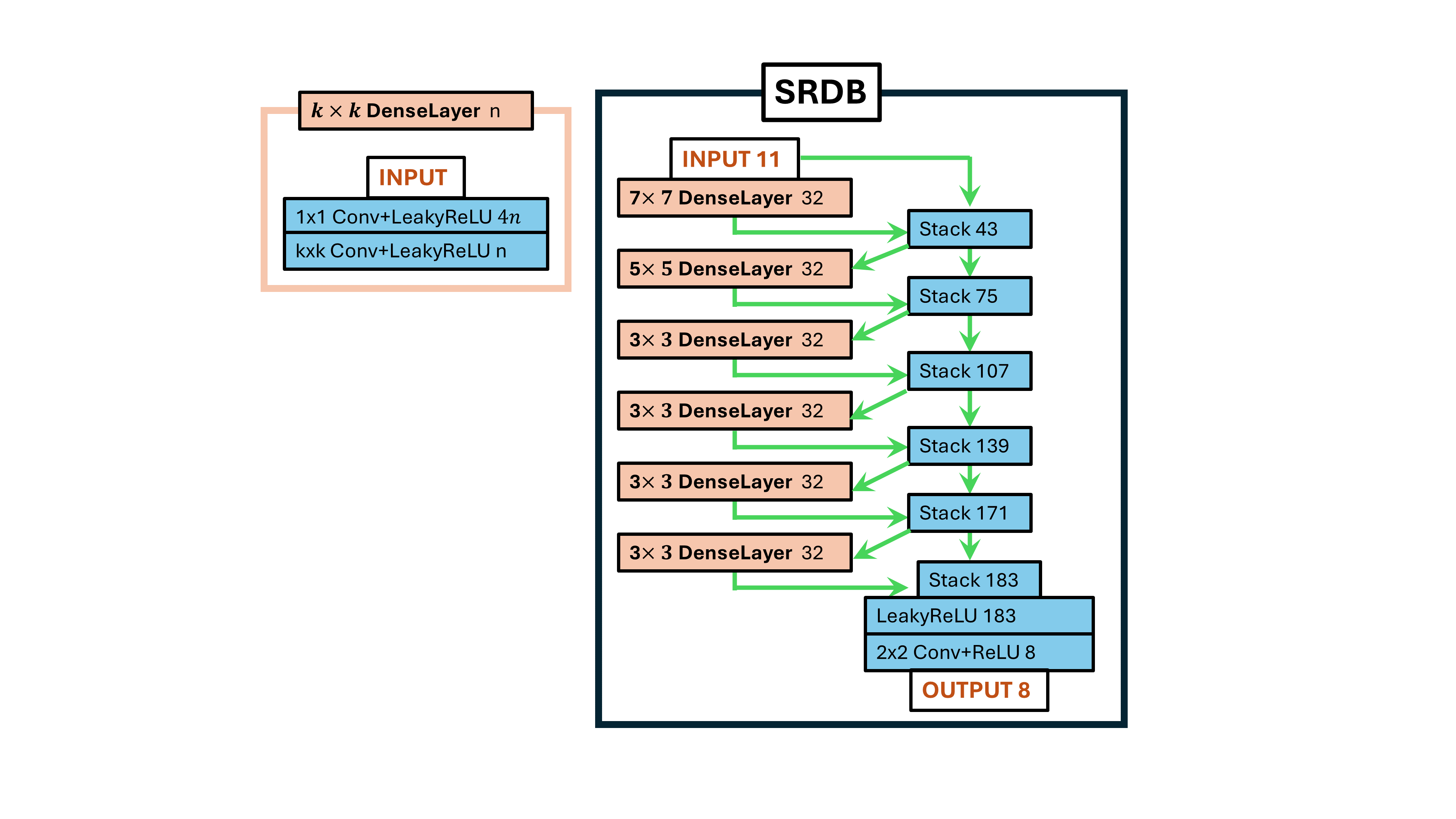}
  \caption{The SRDB network architecture.}
  \label{fig:SRDB}
\end{figure}

\end{appendices}

\end{document}